\documentclass[lettersize,journal]{IEEEtran}
\usepackage{amsmath,amsfonts}
\usepackage{algorithmic}
\usepackage{algorithm}
\usepackage{array}
\usepackage[caption=false,font=normalsize,labelfont=sf,textfont=sf]{subfig}
\usepackage{textcomp}
\usepackage{stfloats}
\usepackage{url}
\usepackage{verbatim}
\usepackage{graphicx}
\usepackage{cite}
\usepackage{amsmath,amsfonts}
\usepackage{algorithmic}
\usepackage{algorithm}
\usepackage{array}
\usepackage[caption=false,font=normalsize,labelfont=sf,textfont=sf]{subfig}
\usepackage{textcomp}
\usepackage{stfloats}
\usepackage{url}
\usepackage{verbatim}
\usepackage{graphicx}
\usepackage{cite}
\usepackage{makecell}

\usepackage{caption}
\usepackage{subcaption}

\begin{document}
	
	\title{Inferring Operator Emotions from a Motion-Controlled Robotic Arm}

	%
	





\author{Xinyu Qi, Zeyu Deng, Shaun Alexander Macdonald, Liying Li, Chen Wang, Muhammad Ali Imran,~\IEEEmembership{Fellow,~IEEE}, Philip G. Zhao,~\IEEEmembership{Senior Member,~IEEE}

\thanks{Xinyu Qi and Muhammad Ali Imran are with the School of Engineering, University of Glasgow,  G12 8QQ Glasgow, United Kingdom. Email: 228889q@student.gla.ac.uk; Muhammad.Imran@glasgow.ac.uk. This work was done during Xinyu's PhD study at the University of Glasgow.}
\thanks{Zeyu Deng and Chen Wang are with Lyle School of Engineering, Southern Methodist University, Dallas, TX 75205, United States. Email: zeyud@smu.edu; cwang6@smu.edu}
\thanks{Shaun Alexander Macdonald and Liying Li are with the School of Computing, University of Glasgow, G12 8QQ Glasgow, United Kingdom. Email: shaun.macdonald@glasgow.ac.uk; liying.li@glasgow.ac.uk.}
\thanks{Philip Zhao is with the University of Manchester, Manchester, United Kingdom. Email: philip.zhao@manchester.ac.uk.}
\thanks{This paper was accepted by the IEEE IoT Journal with minor revisions.}}



\maketitle

\begin{abstract}
A remote robot operator's affective state can significantly impact the resulting robot's motions leading to unexpected consequences, even when the user follows protocol and performs permitted tasks. The recognition of a user operator's affective states in remote robot control scenarios is, however, underexplored. Current emotion recognition methods rely on reading the user's vital signs or body language, but the devices and user participation these measures require would add limitations to remote robot control. We demonstrate that the functional movements of a remote-controlled robotic avatar, which was not designed for emotional expression, can be used to infer the emotional state of the human operator via a machine-learning system.
Specifically, our system achieved 83.3$\%$ accuracy in recognizing the user's emotional state expressed by robot movements, as a result of their hand motions. We discuss the implications of this system on prominent current and future remote robot operation and affective robotic contexts. 
\end{abstract}

\begin{IEEEkeywords}
Emotion Recognition, Robotics, Human-Robot Interaction (HRI), Emotional Movement
\end{IEEEkeywords}



\begin{figure}
\centering
\includegraphics[width=1\linewidth]{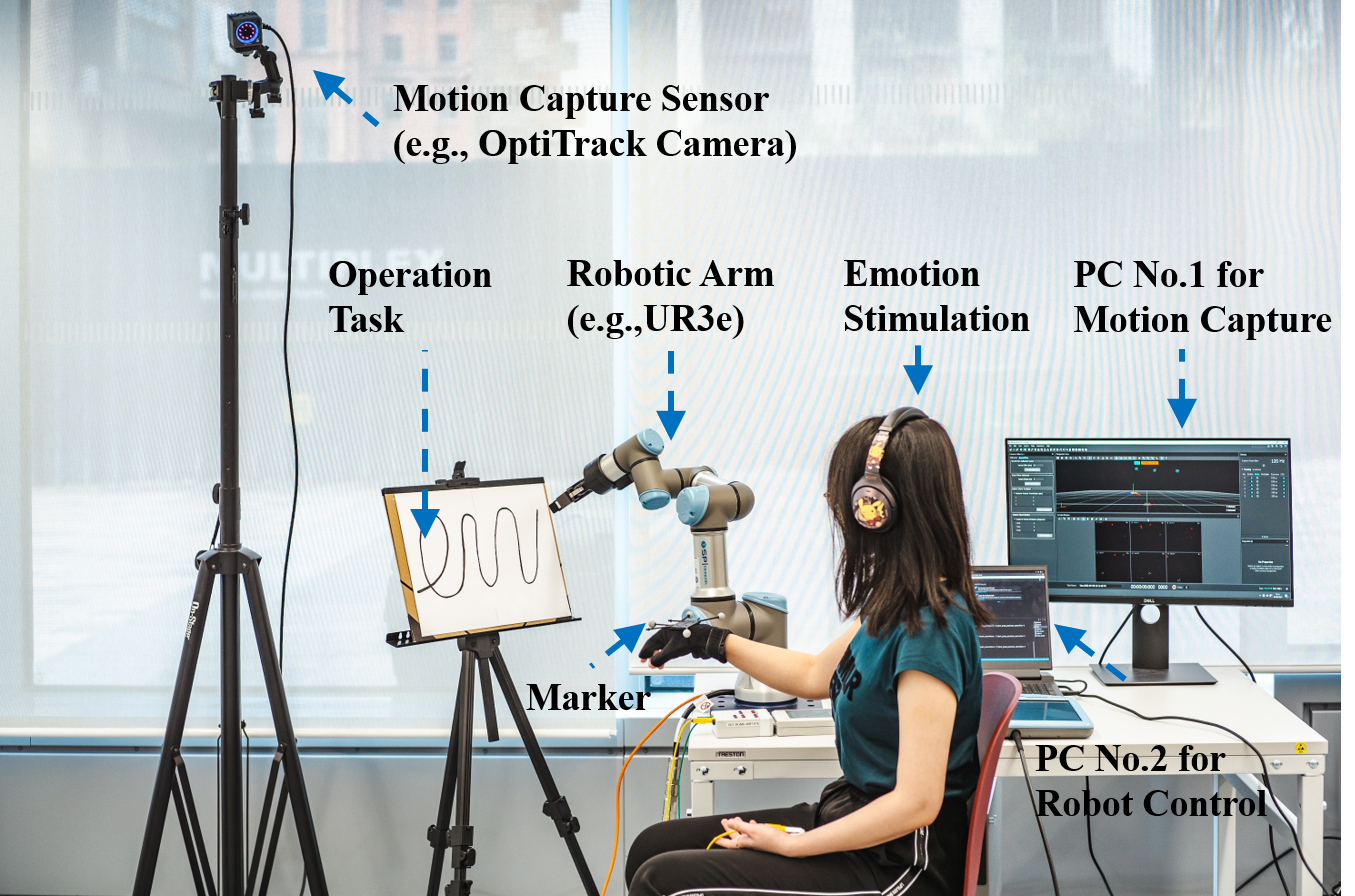} 
\caption{Our robot platform with motion-based emotion transmission.}
\label{fig:pl}
\end{figure}

\section{Introduction}
\IEEEPARstart{T}{} elerobots - robotic avatar systems that replicate the operator's senses and actions - allow operators to overcome physical distance to a remote environment and complete important tasks.
For many safety-critical and precise applications such as telesurgery, nuclear maintenance, and remote driving, understanding the operator's emotional state could be crucial to avoid dangerous outcomes. 
To achieve this, telerobotic systems could be imbued with emotional intelligence systems that allow the inference of these emotional states~\cite{loghmani2017emotional}.
The system could then respond accordingly to prevent or to normalize imprecise movements that result from heightened emotions, facilitating safe and effective interactions~\cite{spezialetti2020emotion}.
In remote driving scenarios, for example, safety measures could be activated when significant fatigue or stress is detected. 
Similarly, such a system could act as a safeguard during telesurgery, where imprecise movements may lead to dangerous outcomes, complementing existing emotion regulation self-assessment procedures~\cite{sharp2020review, nayar2020self}.
Investigating the detection of operator emotional states from telerobotic avatar movement could facilitate safer and more effective interaction in current applications, and inform emotionally intelligent future encounters between humans and telerobotic avatars.
\par
\IEEEpubidadjcol
Existing approaches for robots to classify human emotions rely on analyzing physiological or behavioral data such as facial expressions, speech, or body movement~\cite{8908557}. 
These methods can, however, be unsuitable for telerobotic operation.
Operator movement can interfere with data collection from wearable sensors and physiological electrical signals~\cite{dzedzickis2020human}. 
Meanwhile, facial expressions or voices can be unavailable or occluded in work settings~\cite{khalifa2022face, spezialetti2020emotion}.
Additionally, these methods require users or workplaces to be burdened with additional devices~\cite{mao2015using, dzedzickis2020human}, and the data captured can raise privacy concerns. 
\par
To address these challenges, we propose an alternate approach: inferring the operator's emotions from the movement features of a robotic avatar that it inherits from its operator. 
This method circumvents challenges in existing user-side methods, such as movement artefacts and privacy concerns, while also allowing a better understanding of how affective input impacts the final trajectory of the robot avatar.
This can then be used to design and tune robot-side safety measures.
For example, after observing the resultant changes to a robot's movement based on input from an operator with heightened emotions, an emotive-motion dampening system could be designed to normalize these movements, as discussed in Section~\ref{AI-dampening}. 
Furthermore, observing affective movement on the robot side allows for a single point of affective data collection from many distributed and input devices, creating a system with fewer points of failure and the ability to accommodate different input devices that may not easily facilitate affect recognition in real-time.
\par
Bodily movements can communicate emotional status~\cite{8493586} and research has demonstrated that motion-controlled robots can inherit, or exhibit, unique movement features such as jerk, acceleration, or velocity from the operator's motion. 
Whether operator emotions can be inferred from these inherited motions is, however, unexplored~\cite{huang2021towards}. 
Work by Menck et al.~\cite{menck2023you} found that negative user emotions can be inferred from the movements of a virtual human avatar they control. We verified this finding also applies to telerobotics in a pre-study, then built upon it to develop a \textit{first-of-its-kind} system that infers an operator's emotional state from the movements of the physical robotic arm they control.
We achieved this by utilising a motion-controlled robotic avatar platform and developing learning-based emotion recognition algorithms to analyze the joint and end-effector readings of the avatar's non-stylized motions. 
Inferring emotion by analysing the movements of the robot avatar, as opposed to directly analysing the operator input, confers several potential benefits.
\par
We used two types of telerobotic tasks: 1) mid-air gestures representing industrial~\cite{huang2021robot} and social scenarios~\cite{pollick2001perceiving, loghmani2017emotional}, and 2) a line-tracing task representing safety-critical scenarios~\cite{rakita2020effects}. Five distinct emotional states were elicited from 10 participants using an established affective audio induction method~\cite{kim2008emotion, fakhrhosseini2017affect}, while they controlled the robotic arm to perform tasks. 
Data was collected from both the robotic avatar and an ECG device fitted to the participant, for comparison. We developed a Dynamic Time Warping (DTW)-based algorithm and a Convolutional Neural Networks (CNN)-based algorithm to recognize the user's emotions using either subject-independent or subject-dependent training models. 
Unique features were derived from the robotic arm's movement to infer the operator's emotional state, achieving an average emotion recognition accuracy of 83.3$\%$.
We finish by discussing the implications of our approach on telerobotic applications.  
Our contributions are as follows: 

\begin{figure*}[!t]
\includegraphics[width=\textwidth]{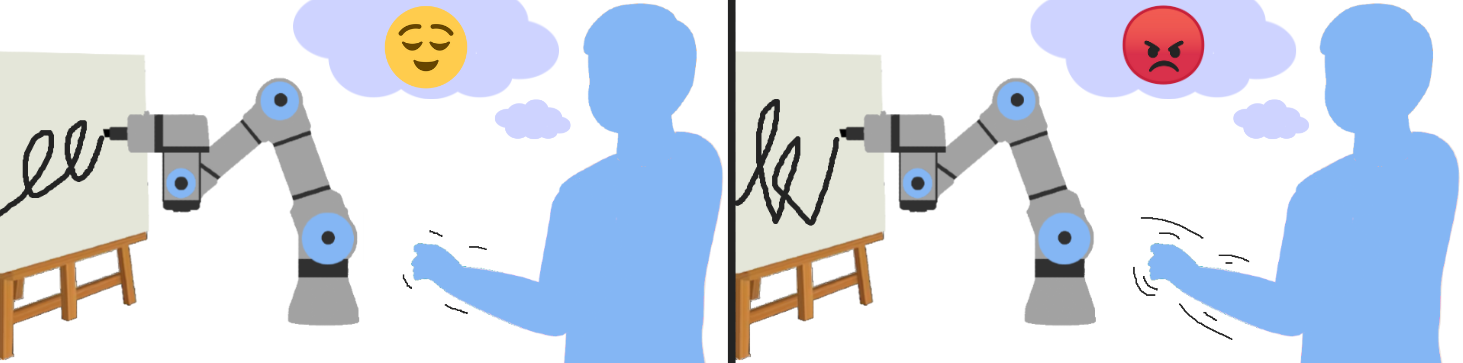}
\caption{The emotions of a human impacting the trajectory of the motion-controlled robot arm they operate. We investigate how the robot's movement can be used to infer these operator emotions.}
\label{fig:teaser}
\end{figure*}

\begin{itemize}
\item We show for the \textit{first time} that user affective states can be accurately inferred from the movements of a physical motion-controlled robotic arm; 

\item We implement two emotion recognition algorithms, based on DTW and CNN respectively, and develop unique emotional features from the robotic avatar's end-effector motions and its joints' spatial and temporal features; 

\item We demonstrate empirically that our approach provides a more suitable alternative for motion-based telerobotic applications than traditional ECG methods;
\item 
We discuss the key implications for current and future Human Robot Interaction (HRI) applications.

\end{itemize}

\section{Background \& Related Work}
\label{Sec:2.1}
\subsection{Modeling Emotion}
\label{subsec:modelemo}

Prior works ~\cite{scherer2000psychological, wang2022systematic} classified the emotion models into four categories, including dimensional, discrete, meaning-oriented, and componential emotion models. Dimensional emotion models include unidimensional and multidimensional models. Russell popularized a multidimensional model\cite{russell1980circumplex, wang2022systematic}, proposing a two-dimensional circumplex emotion model in which the x-axis represents valence and the y-axis represents arousal. 
The x-axis is the pleasure-displeasure axis, where the two ends of this axis represent positive and negative emotions, respectively~\cite{russell1980circumplex}. The y-axis is the activation-deactivation axis, where the two ends of this axis represent high-energy and low-energy emotions, respectively~\cite{russell1980circumplex}. 
For example, joy is a high-arousal and positive emotion, while sadness is a low-arousal negative emotion. Ekman~\cite{ekman1992argument, wang2022systematic} popularized the discrete model and proposed six basic emotions: anger, disgust, fear, happiness, sadness, and surprise.  We chose to use Russell's model in this work~\cite{russell1980circumplex, wang2022systematic}. First, the majority of research on emotion recognition uses the circumplex model, allowing for external validity and comparison ~\cite{dzedzickis2020human, 10.1145/2973750.2973762, kim2008emotion}. Additionally, arousal, as one of the circumplex model's two components, was hypothesised to be especially relevant to telerobotic use, where more active, aroused, and rigorous movement is important to detect. 

In this work, we chose to induce and observe four discrete emotions, one for each quadrant of the circumplex model (Joy, Sadness, Annoyance, Pleasure), with the addition of a neutral emotional state between them. The selected discrete emotions differ significantly in both valence and arousal. More importantly, our selected emotion labels are an established approach used across prior works ~\cite{10.1145/2973750.2973762, soundariya2017eye,  kim2008emotion, loghmani2017emotional, xue2018aspect, zhang2016relieff, dissanayake2019ensemble}.


\subsection{Emotion Recognition during Human-Computer and Human-Robot Interaction}
Emotional intelligence is the ability to recognize and generate emotions~\cite{loghmani2017emotional}. Endowing computers and robots with emotional intelligence could enable more intuitive, efficient, and collaborative human-computer and human-robot interaction~\cite{loghmani2017emotional, 1230215,6662348, spezialetti2020emotion}. By enabling a computer or robotic agent to infer human emotions, they could give corresponding feedback, such as activating alarms and displaying expressive behaviors~\cite{6662348}. Applications of emotion recognition include health monitoring, user experience assessment, intelligent assistance, social interaction~\cite{8908557}, education, surgery ~\cite{kapur2005gesture}, and robot rehabilitation.
Taylor et al. ~\cite{taylor2015using} used skin conductance to detect user frustration in response to system delays. 
In addition, emotion recognition has been shown as a potential aid for people with visual impairments who may find it difficult to interpret others' emotions~\cite{buimer2019opportunities}.
The data utilized for emotion recognition can be categorized into two types: human individual status data and interaction information left on computers or mobile devices~\cite{8908557}. Below, we summarize the existing works that implement emotion recognition using these two types of data. 

\subsection{Emotion Recognition Using Human Individual Status Data}
\label{background:emoreg}
The human individual status data can be further divided into two categories: physiological signals~\cite{dzedzickis2020human} and behavioural signals~\cite{8493586}. 

\subsubsection{Physiological Signals} 
\label{subsubsec:physiological_signals}
Emotion recognition using physiological signals is a hot topic~\cite{shu2018review}. Physiological signals include electroencephalography (EEG), electrocardiography (ECG), HRV, galvanic skin Response (GSR), respiration rate analysis (RR), skin temperature Measurements (SKT), electromyogram (EMG), and electrooculography (EOG)~\cite{dzedzickis2020human}. Among these, EEG and ECG are most frequently used for emotion recognition~\cite{wang2022systematic}. EEG records the electrical activity of the brain by placing electrodes on the head, using 8, 16, or 32 electrode pairs in most cases~\cite{dzedzickis2020human}. ECG detects the electrical activity of the heart by attaching three electrodes around the body~\cite{dzedzickis2020human}, while Zhao \textit{et al}.~\cite{10.1145/2973750.2973762}  proposed a wireless device to capture ECG signals.
There are limitations to these techniques, however.
Human movement produces motion artefacts and interferes with inferring from electrical signals for both EEG~\cite{gorjan2022removal} and ECG~\cite{perez2018main}.
Based on this, it is advised not to collect EEG and ECG data when participants are moving; rather, the subject should be in a calm and stable position~\cite{dzedzickis2020human}, although no prior work has specifically investigated how the impact of telerobotic operation movement on these measures.
\par
Similarly, EMG and EOG, used to detect electrical signals of muscle cells and eye movements, respectively, can be influenced by motion artifacts. SKT is limited by the latency between emotion generation and skin response ~\cite{dzedzickis2020human}.
These limitations provide motivation for an emotion inference system usable in the movement-based scenario of telerobotic operation.
To compare the suitability of our novel approach with an established technique, we utilised ECG in this study.
\subsubsection{Behavioral Signals} Behavioral signals can be divided into two types, verbal signals~\cite{4518767} and non-verbal signals~\cite{8493586}, where non-verbal signals include facial expressions and bodily movements. Voice signals and facial expressions require additional devices and computing resources to process in real-time and are hard to capture while humans are moving. 
Recognizing emotions from gestures and bodily movements remains an under-explored and underestimated topic~\cite{8493586, sapinski2019emotion}.
\par
Emotion-related features can be extracted from \textit{kinematic} features of bodily movement (\textit{e.g.}, head, arm, upper body, or the whole body) and \textit{expressive} features~\cite{5740837}. Kinematic features include velocity, acceleration, and jerk of trajectory~\cite{5740837,bernhardt2007detecting,kapur2005gesture,pollick2001perceiving,loghmani2017emotional}, while expressive features include spatial extent, energy, symmetry, and leaning of the head~\cite{5740837,wallbott1998bodily}. 
Speed is related to how energetic the movement is, acceleration indicates muscle tension, and jerk represents the force~\cite{bernhardt2010emotion}. Prior emotion recognition research has used average hand speed, acceleration, and jerk~\cite{bernhardt2007detecting, ahmed2019emotion}. Others used 14 joint velocities, acceleration, time duration, as well as the mean and standard deviation values of velocity and acceleration~\cite{kapur2005gesture, wang2022systematic}. Pollick \textit{et al}.~\cite{pollick2001perceiving} found correlations between the kinematics features of the arm and emotional states, a finding echoed in many following works~\cite{maret2018identifying, loghmani2017emotional}. In particular, correlations were found between higher arousal levels and several other factors: shorter duration, greater magnitudes of velocity, acceleration, and jerk the movements have. Another correlation was found between positive, higher valence emotions and kinematic features with smaller magnitudes and longer levels of duration.
Prior work~\cite{piana2016adaptive, daoudi2017emotion} has also utilised 3D whole-body motions to extract the kinematic and expressive features mentioned above. Daoudi et al. proposed covariance descriptors to recognize emotions reaching to 71$\%$, comparable with results by human evaluation. 
Piana et al.~\cite{piana2016adaptive} developed a framework that can classify human emotions through their stylized and non-stylized motions, which was used to develop serious games to help autistic children learn to recognize and express emotions by their full-body movement. 
There are, however, privacy concerns when using human facial or movement data directly to infer emotions, as sophisticated camera setups may be required and detailed live video data sent over networks for remote processing.
By instead inferring emotion from a robotic avatar, one could bypass this invasive step.


\subsection{Robot Emotion Expression}

Prior work has explored autonomous robotic emotion expression across different form factors. For example, Ghafurian \textit{et al.}  varied the movements of body parts such as the tail, ears, eyes, and head of the animal-like robot \textit{Miro}~\cite{Ghafurian2022} to express emotion. 
Saerbeck \textit{et al.}~\cite{saerbeck2010perception} explored how a vacuum robot's movement can convey emotion, while others have adjusted the motion parameters of humanoid robots, including acceleration, velocity, and curvature~\cite{xu2013mood, Kaushik2021}.
They showed that robots have the ability to express emotions through their motions, and there exist relations between motion parameters and emotions.
Empowering robots with emotional intelligence could endow robots with the ability to not only recognize emotions but also express emotions. The robot's ability to express emotions can greatly influence the resulting social interaction~\cite{spezialetti2020emotion}. 
Following emotional inference, robots could adjust their emotional display to show empathy or positively influence the emotions of the user. For example, when the user is sad, a robot could attempt to induce happy emotions to comfort them.
While prior work has explored the emotional expression of social robots, we present novel findings on how emotions manifest in the movement of robotic arms used in industrial or medical settings, paving the way for more affective interactions between humans and operated or autonomous robots in current and future human-robot workplaces.

\begin{figure*}
\centering    

\subfloat[Joy.]{  \includegraphics[width=0.24\textwidth]{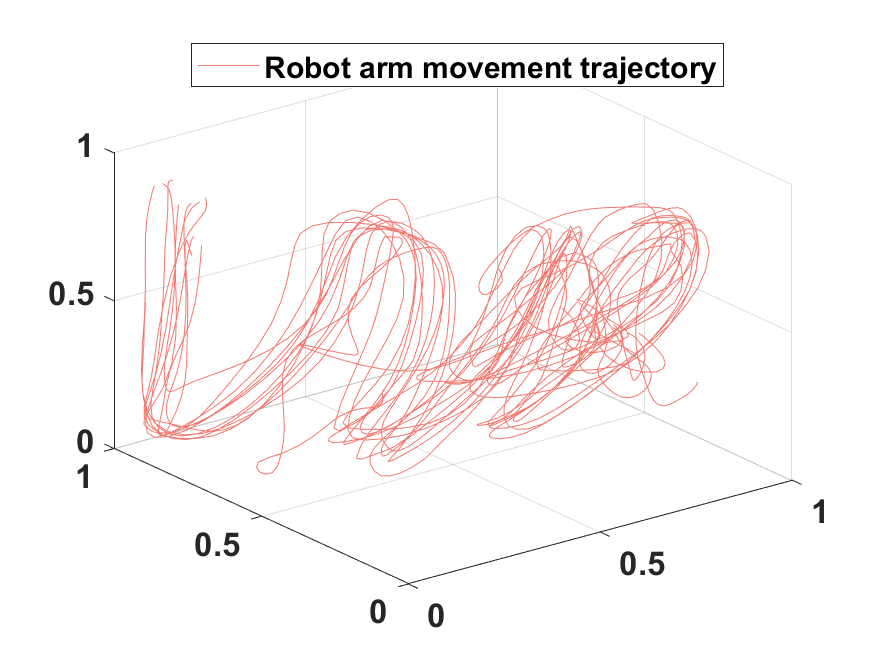}
	\includegraphics[width=0.24\textwidth]{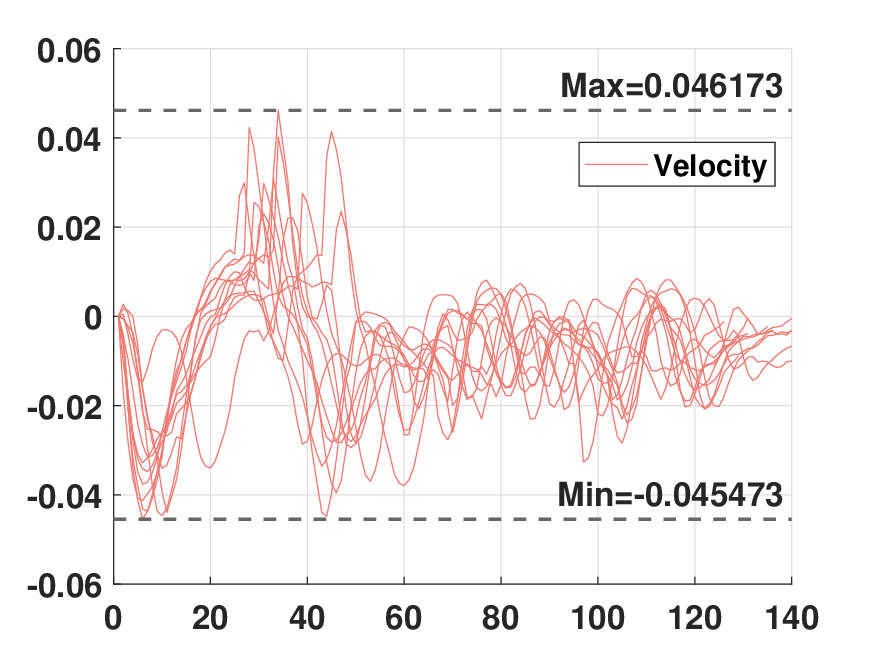}
	\includegraphics[width=0.24\textwidth]{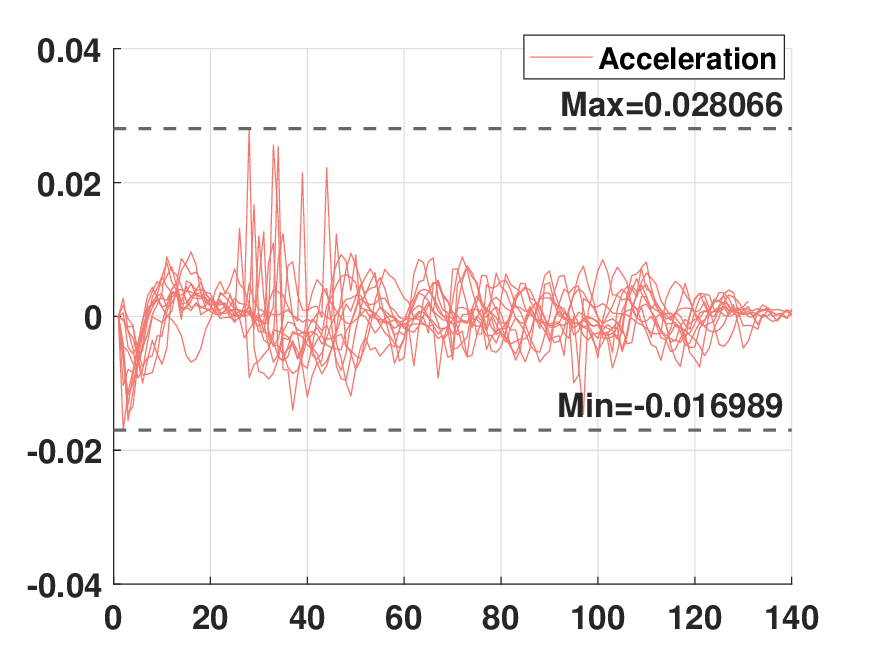}
	\includegraphics[width=0.24\textwidth]{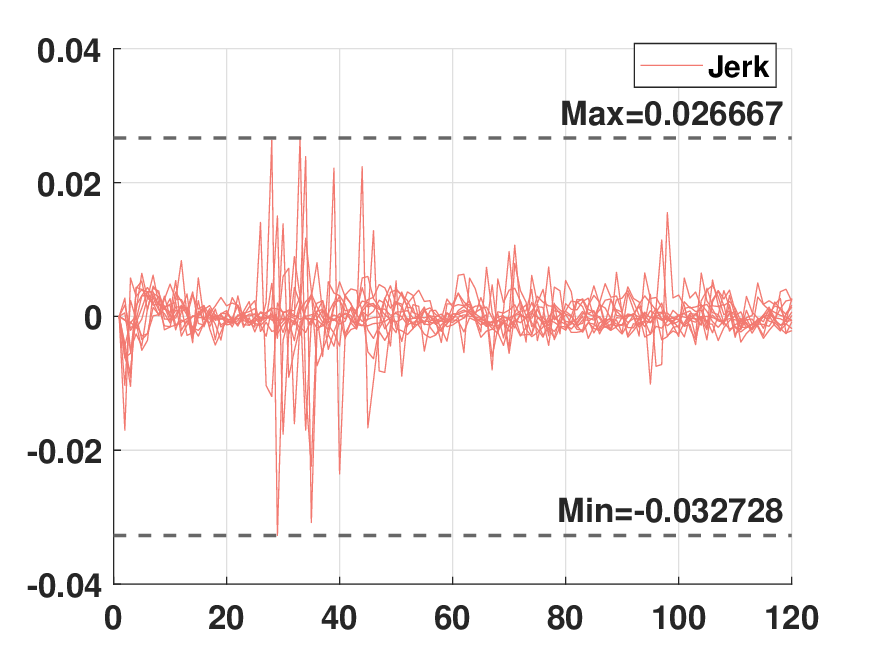}}
\\
\subfloat[Annoyance.]{
	\includegraphics[width=0.24\textwidth]{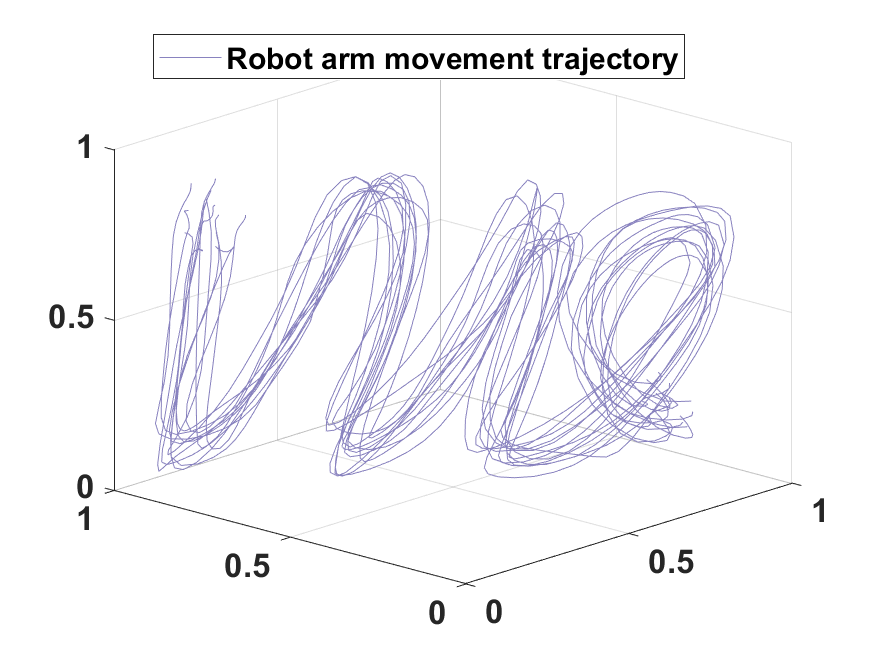}
	\includegraphics[width=0.24\textwidth]{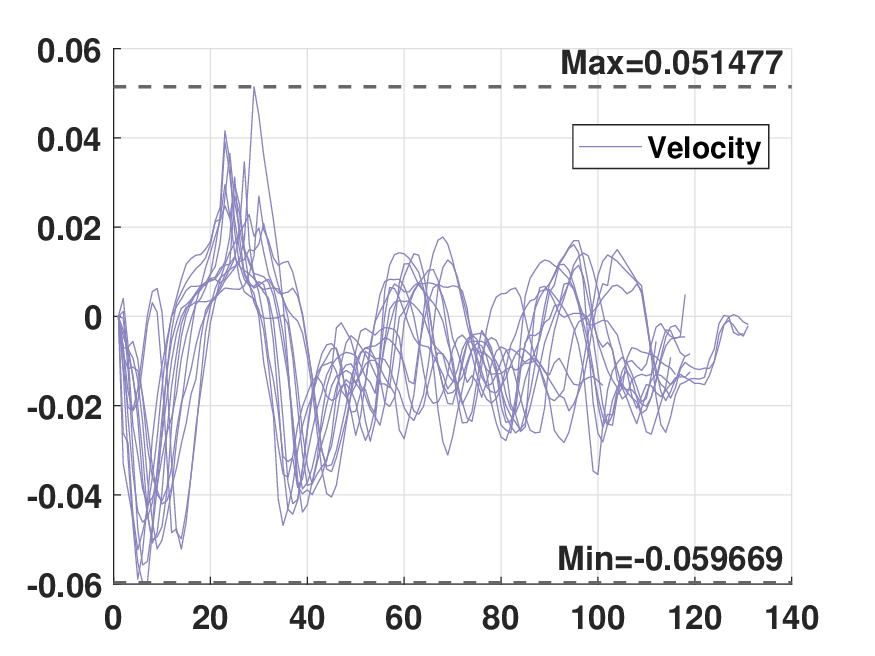}
	\includegraphics[width=0.24\textwidth]{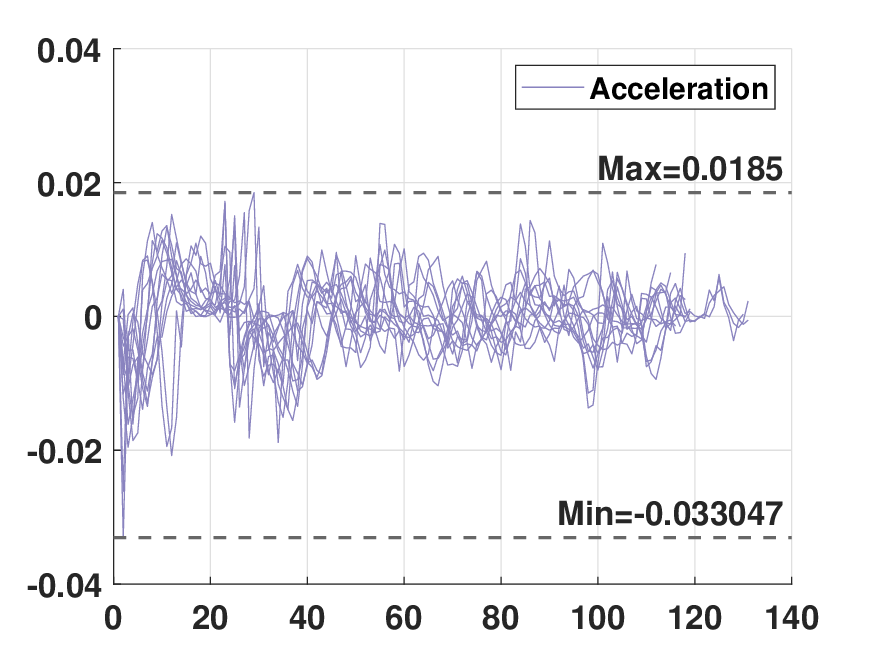}
	\includegraphics[width=0.24\textwidth]{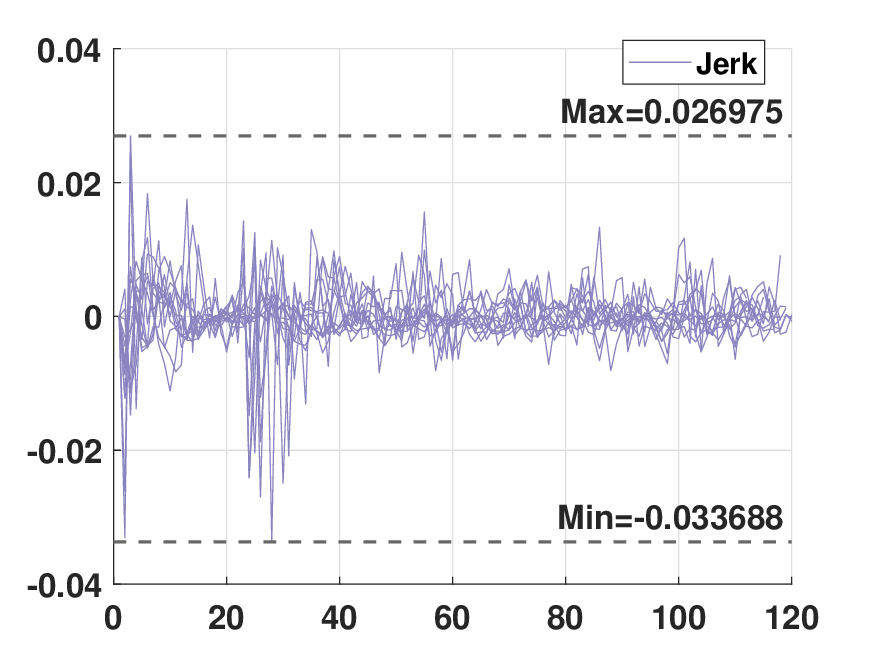}}
\\
\subfloat[Neutral.]{
	\includegraphics[width=0.24\textwidth]{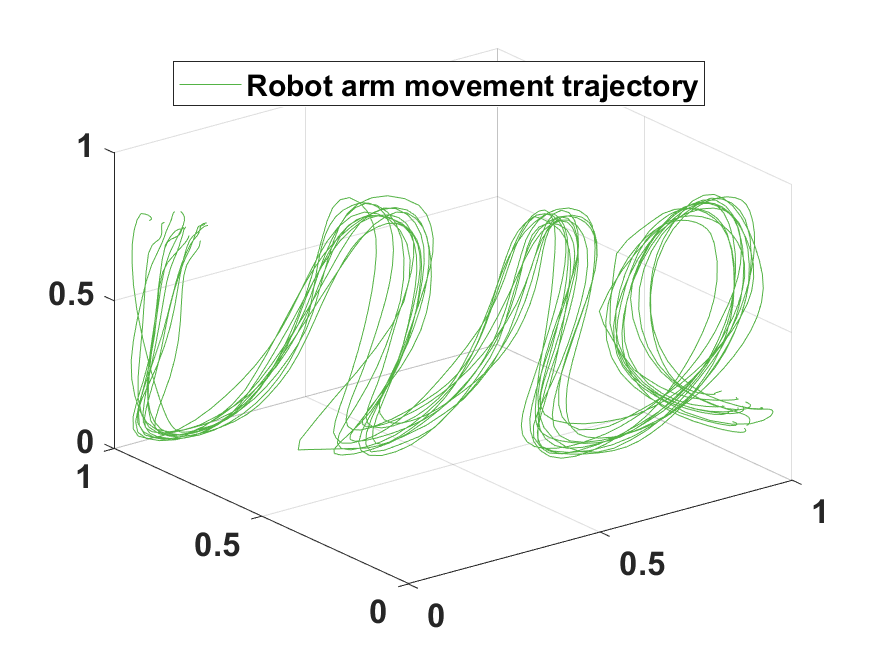}
	\includegraphics[width=0.24\textwidth]{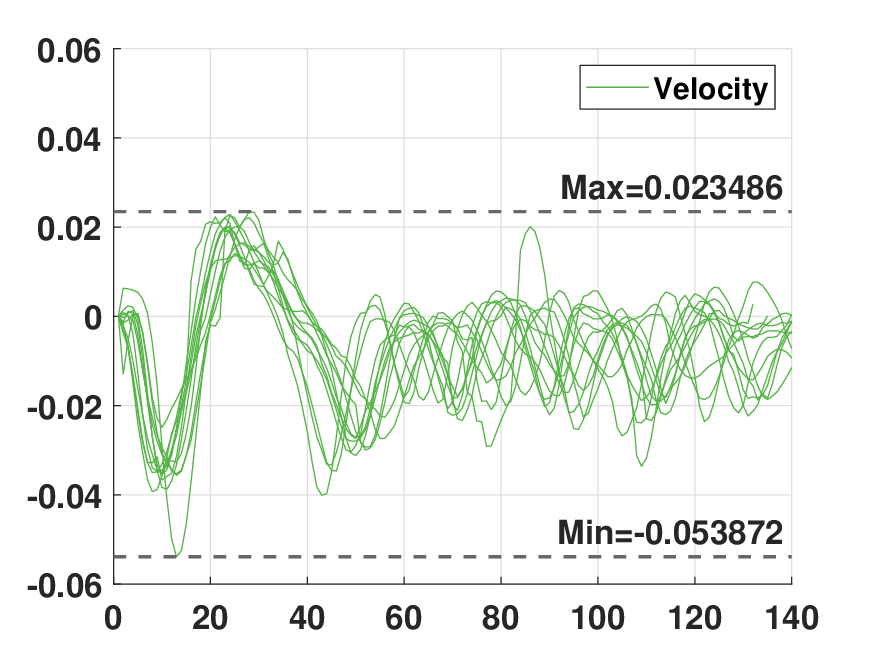}
	\includegraphics[width=0.24\textwidth]{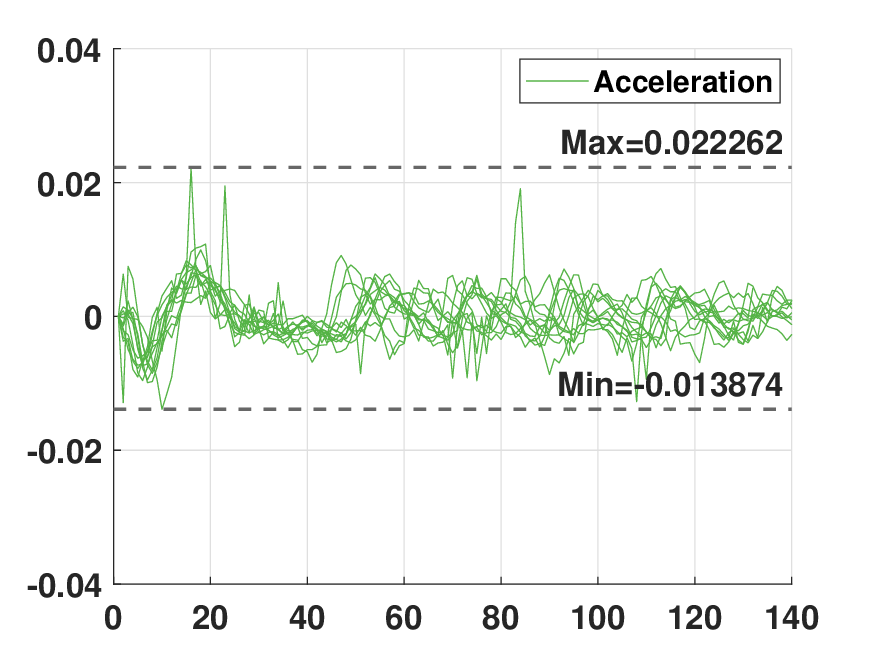}
	\includegraphics[width=0.24\textwidth]{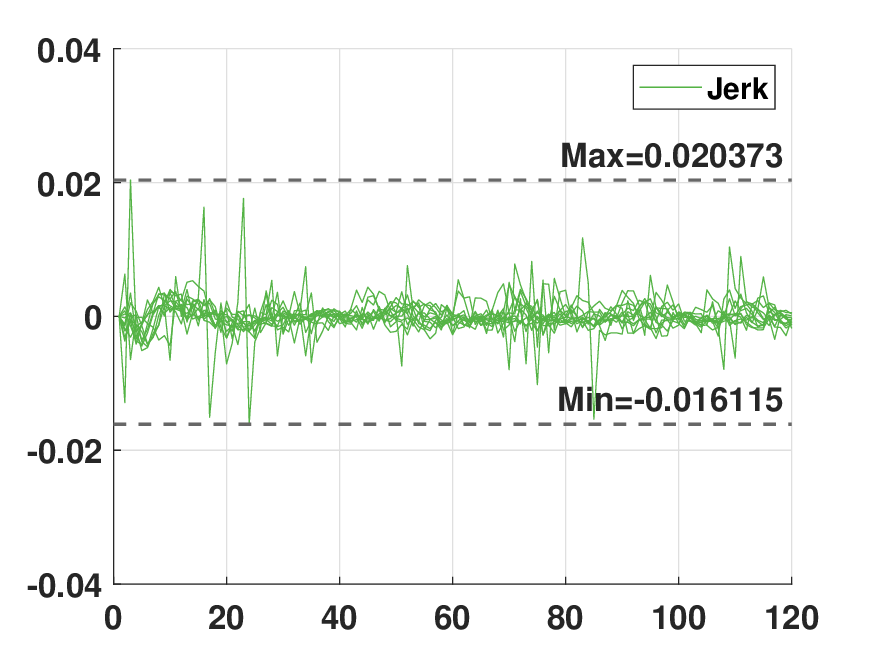}}
\caption{Robot's 3D movement trajectory, velocity, acceleration and jerk plot for joy, annoyance and neutral emotions.}
\label{fig:feature_comparison}
\end{figure*}

\section{System Overview \& Feasibility Pre-Study}

In this section, we introduce a motion-controlled robotic avatar platform and present an initial feasibility study to demonstrate that a robotic avatar can inherit the operator's emotions. Additionally, we introduce the architecture of the proposed emotion classification approach. 
\subsection{Motion-controlled Robotic Avatar Platform}
\label{sec:platform}

We built a motion-controlled robotic avatar platform based on pre-existing control mapping methods \cite{chen2017tracking, walther2017trajectory} where a human interactively controls a robot arm, as shown in Figure~\ref{fig:pl}. An OptiTrack system~\cite{Thornburg03} was built using six cameras arranged in a circle to capture the operator's hand motion trajectory via a glove attached with four markers. 
This system can be used for a teleoperation robot system, where one PC local to the human operator collects their movements, and another PC in a remote area controls the robotic arm. 
A personal computer (PC1) calculated the 3D hand coordinates and sent the pose data to a second PC (PC2). 
PC2 calculated the linear velocity and angular velocity via the received 3D hand trajectory data from OptiTrack and the current robot position data. 
Then PC2 used a Jacobi matrix to calculate the robot’s joint values from the calculated linear and angular velocity.
PC2 was also installed with the robot operating system (ROS) ~\cite{Thornburg04}, which processed the robot joint values and controlled the Universal Robot UR3e~\cite{Thornburg01} in real-time with the help of MoveIt, a 3D motion visualization and control software platform~\cite{Thornburg05}. 
This resulted in a close approximation of the human's trajectory.
During operation, the human receives the visual feedback of the robot's current position and adjusts their hand motions for interactive control, leveraging hand-eye coordination.

\subsection{Pre-Study: Feasibility of Emotion Classification on Robotic Avatar}
\label{subsec:prestudy}







Emotional state impacts our behaviours to varying degrees, and emotional changes can be detected through human motion behaviors~\cite{911197, zhang2016multimodal}.
Are these changes inherited by the robotic avatar? Huang \textit{et al}.~\cite{huang2021towards} showed that a robot could inherit the human operator's behaviours and human behavioural biometrics. It was unclear at this stage, however, if the affective state is also inherited. 
\par
To assess if this approach may be viable for further exploration, we conducted a pre-study that investigated whether the robotic avatar could inherit affective states experienced by a human operator's emotions, such that it manifests in distinct differences in motion. 

One volunteer was employed. During the experiment, the participant was asked to listen to audio files to stimulate emotions. The audio files were picked up by the participant in advance. This audio induction method is a well-established approach, including the use of music~\cite{kim2008emotion} and sound~\cite{ fakhrhosseini2017affect}. 
In the pre-study, we used two opposing emotions (joy and annoyance) from the set of four emotions from each quadrant of the circumplex model that we later employed in the main study Section~\ref{subsec:modelemo}, as we explored a more focused proof of concept.
We chose this pair of opposing emotional states as we hypothesised they were most likely to exhibit different movements in the robotic arm, as well as a neutral baseline state.
The participant was asked to draw the task ``Lw'' according to a reference trajectory on a whiteboard while controlling the robot. He repeated drawing the task 12 times for each emotion, in 3 total blocks (one per emotion) for a total of 36 repetitions. The participant performed all 12 repetitions for each emotion one after the other. Then, we collected the robot arm's end-effector trajectory data and extracted three features (velocity, acceleration and jerk) from the trajectory and then visualised them. 
Figure~\ref{fig:feature_comparison} shows 4 factors of the robotic arm movement including trajectory, velocity, acceleration, and jerk plots for three different emotion conditions. In each figure, every single line represents one single repetition. The illustrated figures show the difference in the robot's motion information between different emotions.

We observed that, when the operator was emotionally neutral, the robotic avatar's trajectories were less frenetic or dramatic, while the joyful and annoyed trajectories featured more sudden shifts in motion, as well as greater peaks and troughs. 
More specifically, the widest range of trajectory velocities occurred during annoyed emotional states, followed by joyful states, with neutral states having the smallest range.
Trajectory acceleration was more consistent during neutral states than joyful and had fewer fluctuations than during annoyed states. 
Joyful and annoyed trajectories were more disordered and erratic than neutral and jerks were more common. 
This indicated the operator was either less concentrated on their motions when influenced by these stronger emotions~\cite{taylor2005interaction, mitchell2023emotion}, or that these high arousal emotional states~\cite{russell1980circumplex, wang2022systematic} caused with the operator to subconsciously performing stronger, more active and more erratic gestures, an effect identified by Glowinski \textit{et al.} \cite{5740837}, Pollick \textit{et al.} ~\cite{pollick2001perceiving} and Wallbott\textit{et al.}~\cite{wallbott1998bodily}.
These effects can also be observed in mouse movement and touchscreen dynamics~\cite{yang2021review, kolakowska2013review}.
Initial findings from this pre-study suggested that when operators express higher arousal emotional states, their motions become more energetic and less stable, further suggesting that emotional state can influence a robotic avatar's trajectories. This pre-study was the first evidence that robot arms' movements meaningfully change based on user emotions.

\subsection{System Architecture}
\begin{figure}
\centering
\includegraphics[width=\linewidth]{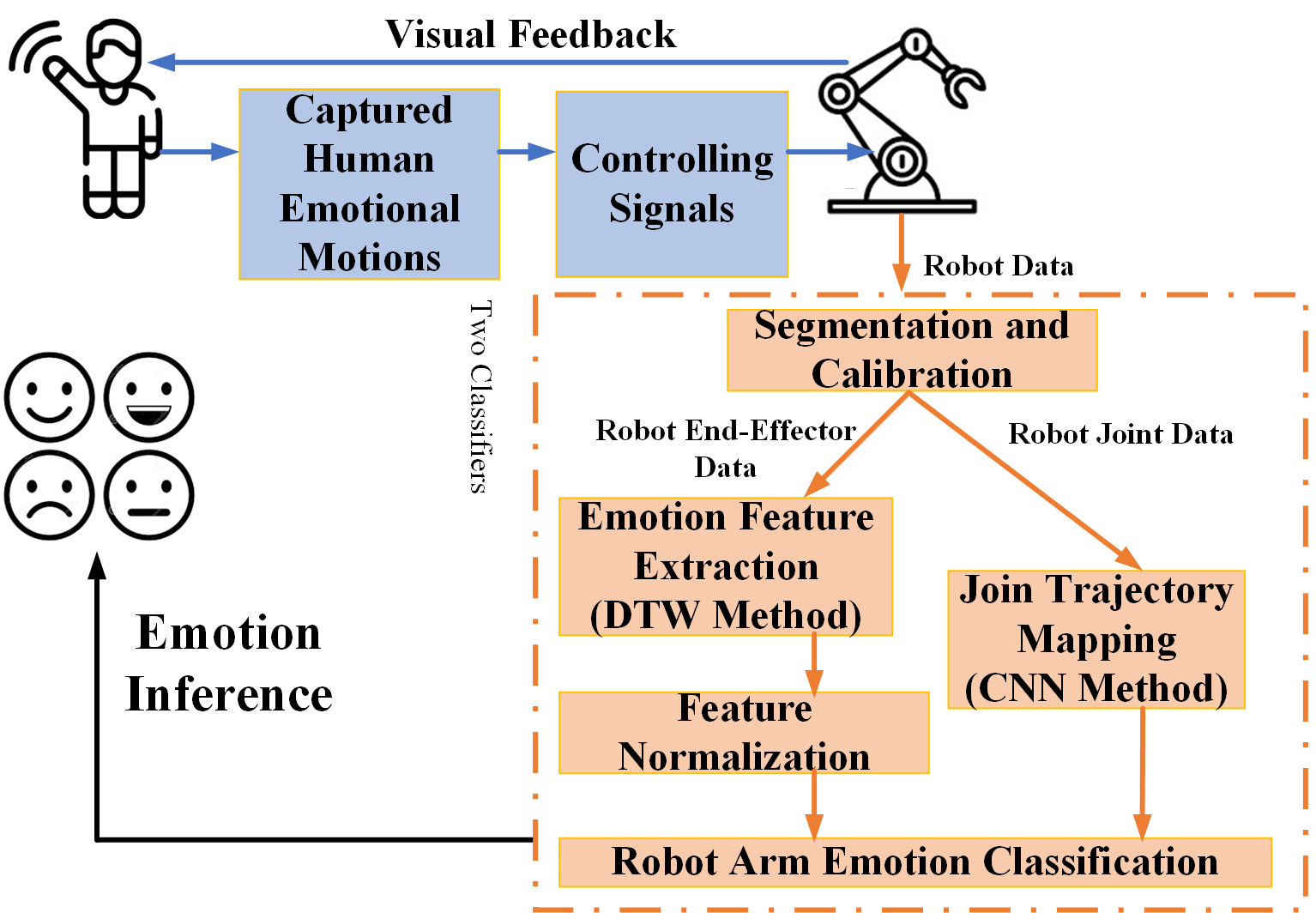}
\caption{The architecture of the proposed human emotion inference through the robotic avatar.  }
\label{fig:attchitecture}
\vspace{-2mm}
\end{figure}

The basic aim of our system is to use the robotic avatar's motions to infer the operator's emotions during interactive control. The architecture of the proposed classification approach is illustrated in Figure~\ref{fig:attchitecture}. An operator controls the robot to perform motion tasks while in different emotional states. The operator's hand motions are first captured by a motion capture system and further calculated to control command sequences that are sent to the robot to enable the robot to execute tasks in real-time. Meanwhile, the operator observes the robotic avatar's motions and adjusts his/her motions to perform interactive control, using hand-eye coordination. 
Two emotion classification methods are deployed on the robotic avatar. When the robotic avatar executes the motion tasks, the values of the robot's joints and the end-effector data (the endpoint movement of the robotic arm) are input to our classifiers. 

\noindent\textit{Segmentation and Calibration} is first applied to acquire the instances of the performed motion tasks. To observe how the operator's emotion influences the robotic avatar's motions, we developed two emotion classification algorithms, a DTW-based algorithm, and a CNN-based algorithm. 
In the DTW-based algorithm, segmented end-effector trajectory is used, after which \textit{Emotion Related Feature Extraction} derives unique motion features to capture the operator's emotion information. 
The derived features are then normalized and analyzed by DTW to infer the operator's emotion. 
In the CNN-based algorithm, the segmented time series of all the robot's joint rotation angles are analyzed and mapped into polar coordinates by \textit{Joint Trajectory Mapping} to generate colour gradient polar plots, with different colours to present different joints and a light-to-dark gradient to present time (see Figure~\ref{fig:joint}). This approach presents both spatial and temporal features of a task instance as a 2D image.
These colour gradient plots are evenly split into training and testing datasets and fed into a CNN-based model for emotion classification. 
Finally, based on the classification result, we can infer the operator's emotions. 
At this stage, a real-world system could decide whether to continue or abort the operation according to the inferred emotions and the importance of the current task.

\section{Emotion Classification Algorithm Design}

In this section, we introduce the adopted emotion model and give detailed descriptions of methods for feature selection and emotion classification.  

\subsection{Dimensional model of emotions}
\label{subsec: Dimensional model of emotions}
\begin{figure}
\centering
\includegraphics[width=\linewidth]{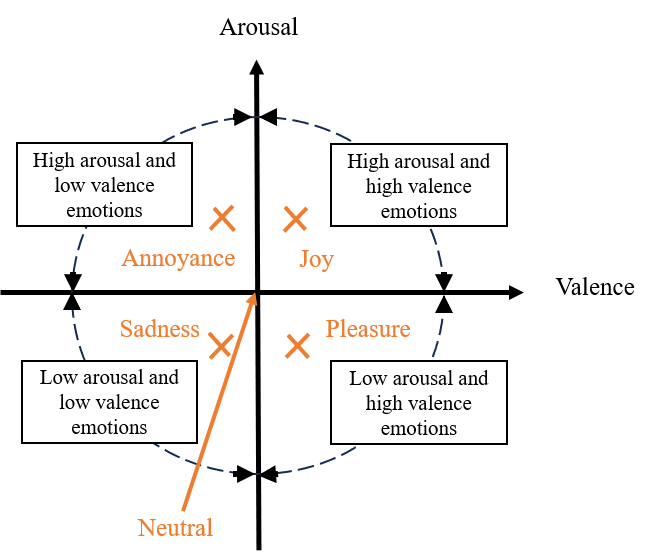}
\caption{The four emotional states we induced in this work, Joy, Pleasure, Annoyance, and Sadness mapped to each of the four quadrants of Russel's circumplex model of emotion, with neutral at the origin.}
\label{fig:emotion model}
\end{figure}

Our work utilised the two-dimensional emotion model established by Russell \textit{et al.}  ~\cite{RUSSELL1977273}, a common approach for researchers ~\cite{kim2008emotion, 10.1145/2973750.2973762, dzedzickis2020human, 10.1145/3536221.3556572} (see Section ~\ref{Sec:2.1}). The x-axis represents the valence, and the y-axis represents the arousal of the emotion. We used five basic emotions defined in each of the four quadrants of the model, respectively: joy, pleasure, sadness, annoyance, and a central neutral emotion~\cite{10.1145/2973750.2973762, loghmani2017emotional, kim2008emotion}, as shown in Figure~\ref{fig:emotion model}.
Joy is categorized in the positive valence and high arousal quadrant, while pleasure is categorized in the positive valence and low arousal quadrant. Annoyance is categorized in the negative valence and high arousal quadrant, while sadness is categorized in the negative valence and low arousal quadrant. 
We collected five basic emotions that were used across many prior works and the years in the field of affective computing, as explained in Section~\ref{subsec:modelemo}. In the future, we can analyse non-basic emotions, such as confusion, frustration, boredom, and engagement, which may occur when participants interact with robotic interfaces in the real world~\cite{d2013beyond,bosch2015automatic}. Future work could also build on our work to test how well it can apply to real HRI contexts, such as medicine and education, as in different contexts, the requirements and benchmarks for emotion recognition may change. For example, in telesurgery, emotion recognition should focus on the intense emotional status, while in e-learning, emotion recognition may focus on more positive and negative emotion detection.

\subsection {Data Segmentation, Normalization, and Calibration}
The robot's positional data was first segmented into instances based on the end-effector's trajectory. We set a threshold to the velocity of the end-effector's trajectory and use this threshold to determine each instance's starting and ending positions. Each repetition of one completed task is regarded as one instance. The x, y, and z axes of segmented instances were normalized into a $1 \times 1 \times 1$ bounding box to make them comparable. In addition, the starting position of the trajectories was aligned with the origin of the UR3e to make the instances spatially comparable. 

\subsection{Robotic Avatar Emotion Classification by Using DTW}
\label{subsec:dtw}
We developed a DTW-based algorithm to classify the robotic avatar-inherited human emotions, utilizing positional data of the robot end-effector within the segmented instances.

\begin{figure}[t]

\centering
\includegraphics[width=\linewidth]{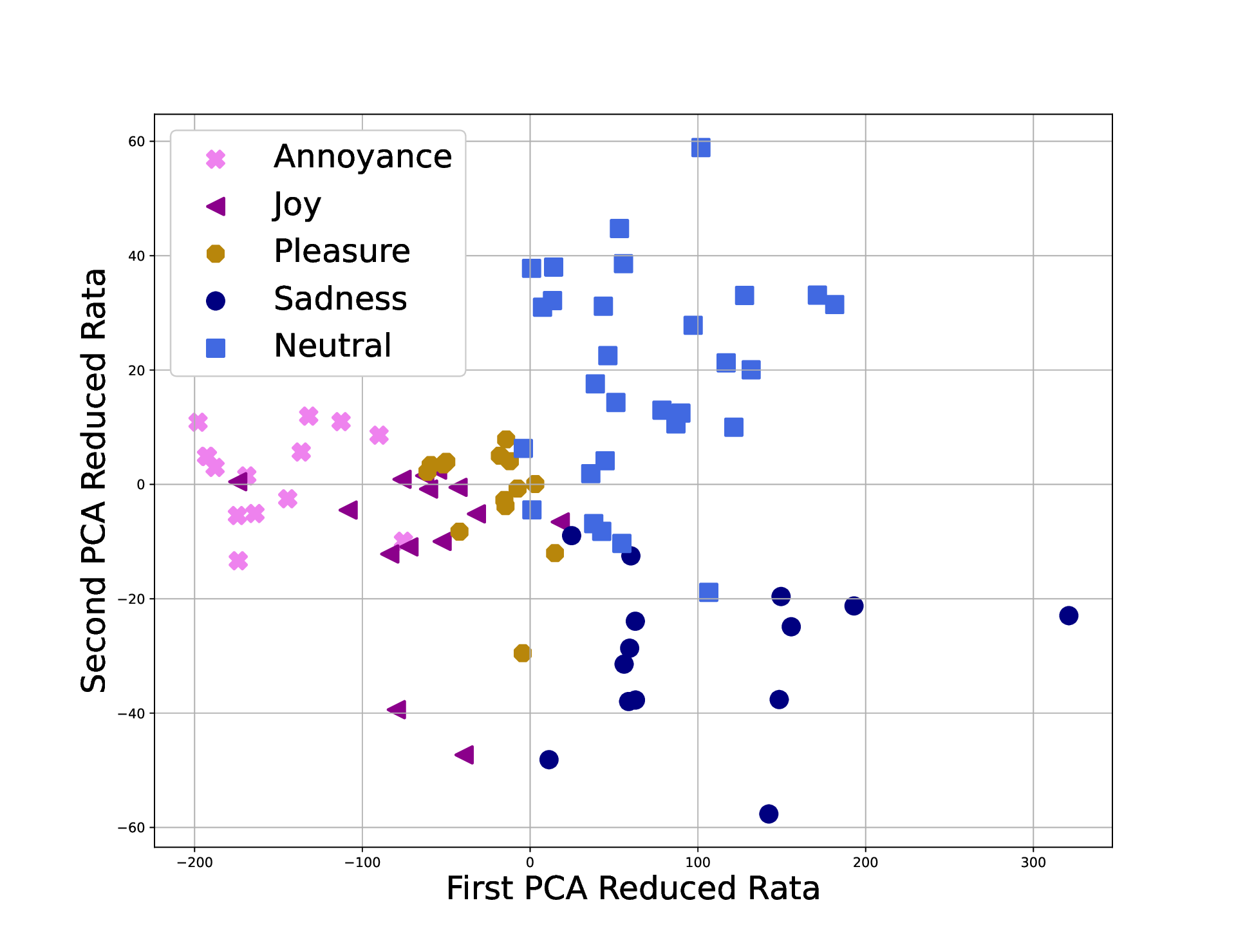}  
\caption{Emotion distributions of one subject's subject-dependent data by using PCA reduced features.}
\label{fig:pca}
\end{figure}

\subsubsection{Emotional Related Feature Extraction}
To capture the operator's emotions manifesting in the robot motions, 20 emotion-related features were extracted from end-effector data and time-sequenced. 
These features were categorized into two types: kinematic features (3D trajectories, 3D velocity, 3D acceleration, 3D jerkiness, position difference) and expressive features (slope angle, curvature, energy, spatial extent, time range). 
We chose to examine motion features to further establish whether distinct differences in trajectory can be observed between a wider range of emotional states, following promising early results from our pre-study (see Figure~\ref{fig:feature_comparison}). 
It was natural to investigate the expressive features, including energy, spatial extent, and time duration, as these have been shown in prior work to convey emotional information in other contexts~\cite{5740837}. We calculated the energy of trajectories by calculating the entropy of signals and the spatial extent of each task instance by calculating the size of each trajectory. Higher energy motion relates to high arousal emotion, while lower energy relates to low arousal energy~\cite{wallbott1998bodily, wang2022systematic, 5740837}. The motions' use of space indicates valence of emotions~\cite{wallbott1998bodily, wang2022systematic, 5740837}. The time range is a key factor in judging the human emotion's arousal level~\cite{7807324}, so we calculated the time of performing each trajectory. In addition, we extracted jerkiness, slope angle, and the curvature of trajectories to represent the motion smoothness, as Glowinski \textit{et al.} ~\cite{5740837} showed that smoother movement correlated with emotional expressions. 

Principal Component Analysis (PCA) is used for dimension reduction while preserving the maximal information ~\cite{arora2018facial}. It was applied to these features to validate whether they can be used to distinguish different emotions. Specifically, we calculated the mean, variance, and standard deviation values for each feature sequence, resulting in 39 static feature values in total for each instance. Then we reduced this 1$\*$39 vector to a 1$\*$2 vector using PCA and visualized each instance according to different emotions. We applied this method to all participants, and they showed similar boundaries between different emotions. Figure~\ref{fig:pca} shows an example of one subject's instances in five affective states, and different emotions are represented by different colors. Clear and discriminated boundaries can be observed between all the different emotions, indicating that the features we extracted can be used to distinguish between them.

\begin{figure}[t]
\centering
\includegraphics[width=.65\linewidth]{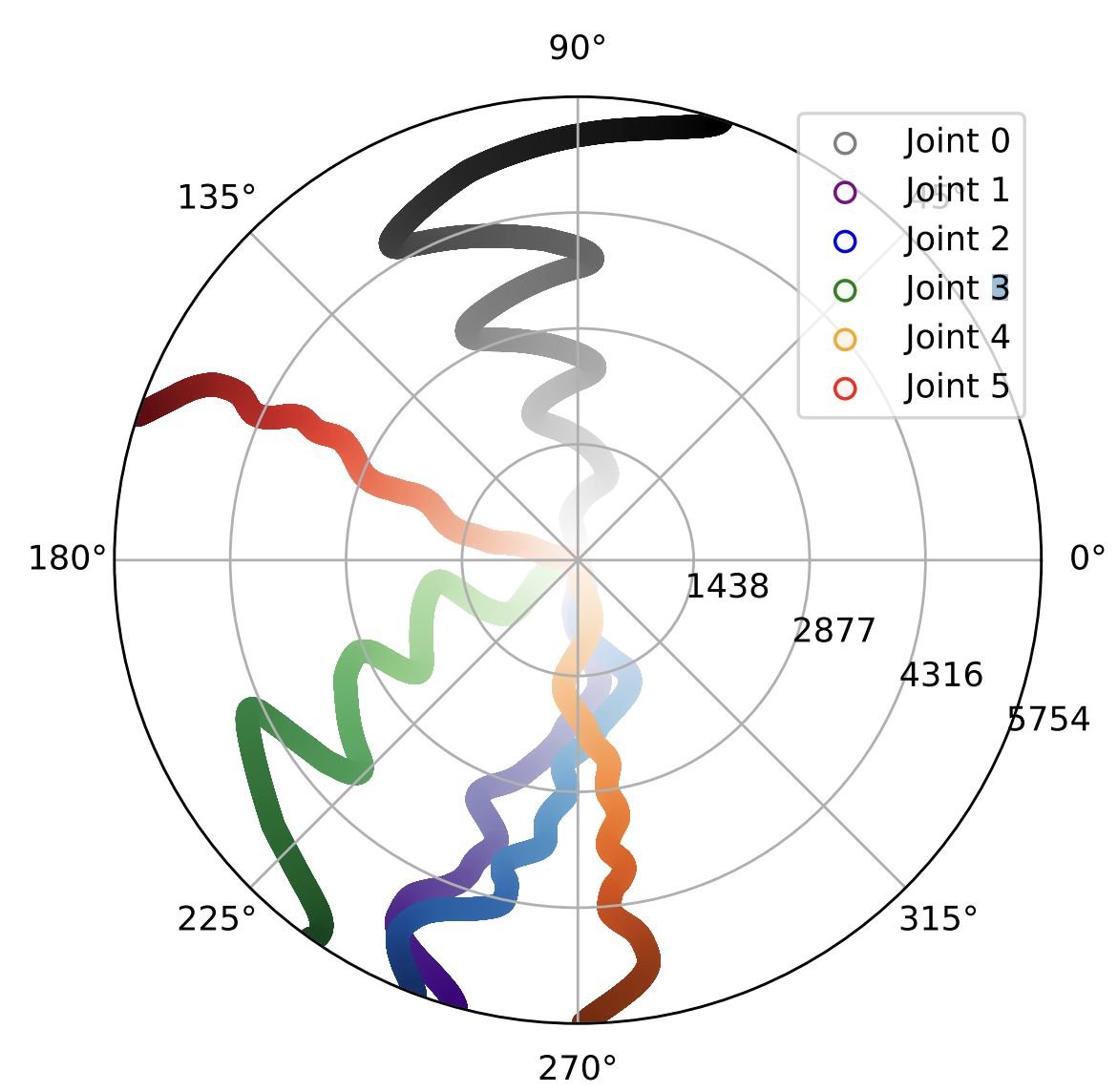}
\caption{Joint rotation angular trajectory mapping.}
\label{fig:joint}
\end{figure}

\subsubsection{DTW-based Emotion classification}
The extracted features were normalized to make the data comparable, then fed into a DTW-based algorithm for emotion classification. DTW allows non-rigid warping along the temporal axis and can also compensate for the feature difference caused by different motion speed~\cite{tian2013kinwrite}. Furthermore, DTW requires less training data and computational resources than other learning-based algorithms and has been used in related work before, as Taghavi et al.~\cite{taghavi2021online} effectively used DTW to facilitate the mimicry of human motions by the human-like robot NAO. Our DTW-based algorithm first selects templates for each emotion from the training data, then compares the testing instance to those templates. The instance is classified to the emotion templates with it has the shortest DTW distance. Specifically, each emotion's templates are the most representative instances of this emotion. They are selected by a pairwise comparison within each emotion during training. For every two instances, DTW distance is calculated between their feature sequences. The templates are selected for each instance based on whose DTW distances to all the other instances within the training dataset are minimal. The template number selected for each emotion can be tuned according to the number of users involved. We used 5 emotion templates in total to achieve both high performance and low computational cost. 

\subsection{Robotic Avatar Emotion Classification by Using CNN}
\label{subsec:cnn}

We also developed a CNN-based algorithm to classify emotions by analyzing the robotic arm's joint data. Specifically, the joint rotation time series is mapped into polar coordinates, where the polar degree represents the joint rotation angles and the radius represents the time frames. As illustrated in Figure~\ref{fig:joint}, each joint of the robotic arm is plotted as a curved line to show the spatial information, with a colour gradient from light to dark indicating temporal information. The polar plots were scaled to a 150 $\times$ 150 resolution image with no background grid before being input to the CNN.

This conversion allows the model to learn from the relationships and dependencies between data points, enabling it to capture complex patterns that would otherwise be lost if the data were simply flattened or represented as a regular grid. We used a CNN because it can effectively learn spatial relationships and local dependencies from 2D image representations. The image-based approach allows the CNN to capture joint coordination in a simpler and more interpretable way. Similar time-series-to-image conversions have been used in related works such as motion \cite{deng2023enhanced} and audio \cite{hershey2017cnn}.

\begin{figure}[t]
\centering
\includegraphics[width=\linewidth]{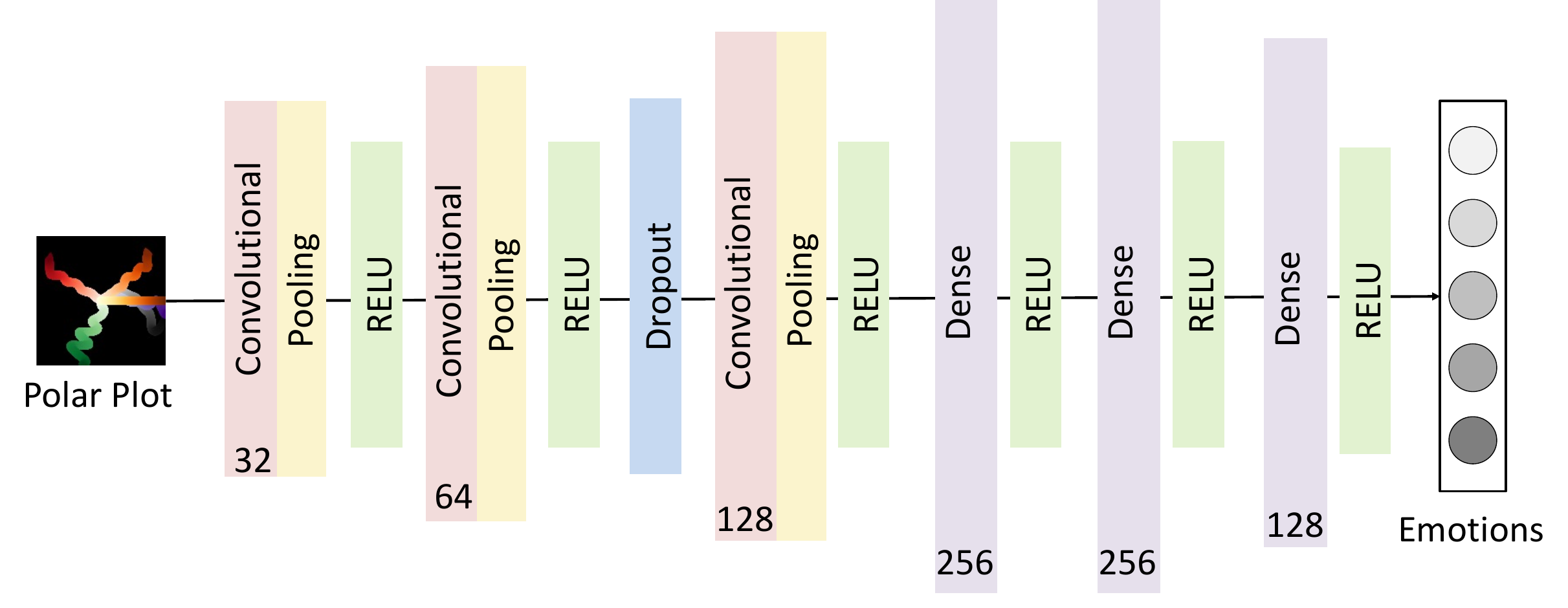}  
\caption{The CNN architecture. }
\label{fig:CNN}

\end{figure}

The structure of the CNN is built from three convolutional layers followed by four fully connected layers, their size (number of neurons) and connections are illustrated in Figure~\ref{fig:CNN}. The input layer is 150$\times$150$\times$3. The output channels of each convolutional layer are 32, 64, and 128, respectively, and the kernel size is all 3. Each convolutional layer is followed by a ReLU activation and a max pooling layer. A dropout layer (rate = 0.2) is applied after the second convolutional layer to prevent overfitting. The size of the final dense layer has the same output size as the number of emotions. The extracted feature maps are then flattened and passed through four dense layers, where the final layer outputs the number of emotion classes. Convolutional layers are efficient in extracting high-level features in the input image, and the dense layers flatten the features and make classification decisions.
The network is trained using the Adam optimizer with an initial learning rate of 1e$^{-4}$. The batch size is 8, and training is run for 100 epochs. The model's FLOPs is 0.16G and Params is 0.85M.

\subsection{ECG Signals Emotion Classification}
\label{subsec:ECG ER}
ECG has potential suitability for our experiment when compared to other options, such as facial or voice. Changes in facial expressions or voice signals may be hard to capture in a telerobotic scenario, as this would require head-facing cameras to be installed in telerobotic workstations, while speech is not inherent to many tasks. 
In addition, our selected emotions can be effectively detected by electrocardiogram. Existing work~\cite{ zhao2016emotion, hsu2017automatic} has proven to use electrocardiograms to detect these emotions, including joy, sadness, pleasure, and annoyance, reaching emotion recognition rates of 90.0$\%$ ~\cite{hasnul2021electrocardiogram}. 
ECG as an example of a prevalent and widely used technique that may struggle to be suitable in telerobotic scenarios, due to short interaction times and movement artefacts. In this work, it acts as a baseline emotion recognition measure to explore if our approach could offer an alternative that is better suited to telerobotic operations.

An ECG detects the electrical activity of the robot operator's heart in real time~\cite{dzedzickis2020human}. Each heartbeat is a specific waveform, QRS complex waveform, caused by the ventricles' contraction, where there are 6 main points (P, Q, R, S, T, U) ~\cite{dzedzickis2020human, 10.1145/2973750.2973762, kim2008emotion}.  
We then extracted emotion features from time-domain and frequency-domain features, including pNN50, Welch PSD: LF/HF, Lomb-Scargle PSD: LF/HF, Autoregressive LF/HF, Poincar SD1, Poincar SD1/SD2, and Detrended Fluctuation Analysis(DFA), following well-established methodology from across prior work~\cite{10.1145/2973750.2973762, dzedzickis2020human, kim2008emotion, hsu2017automatic}.
Then these features are fed to the Support Vector Machine (SVM)~\cite{10.1145/2973750.2973762} for emotion classification.

\begin{figure*}[!t]
\centering
\subfloat[``S'']{\includegraphics[width=0.22\textwidth]{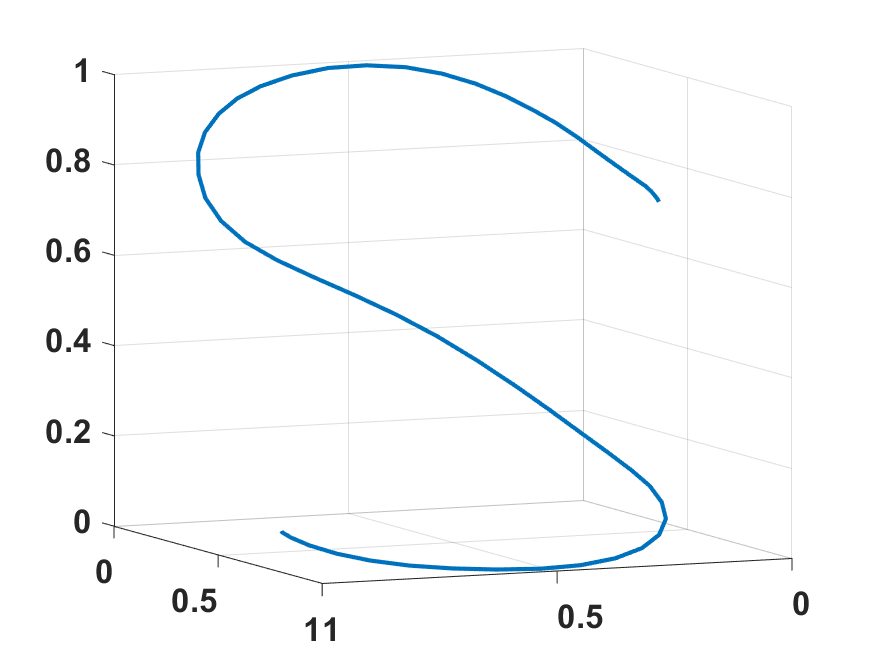}}    
\subfloat[``Star'']{\includegraphics[width=0.22\textwidth]{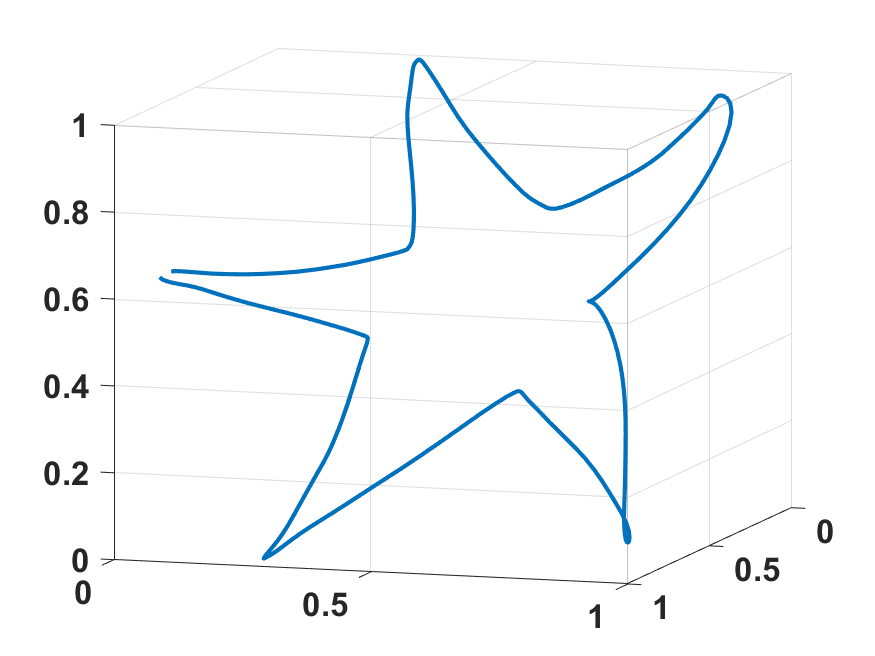}}
\subfloat[``Stir'']{\includegraphics[width=0.22\textwidth]{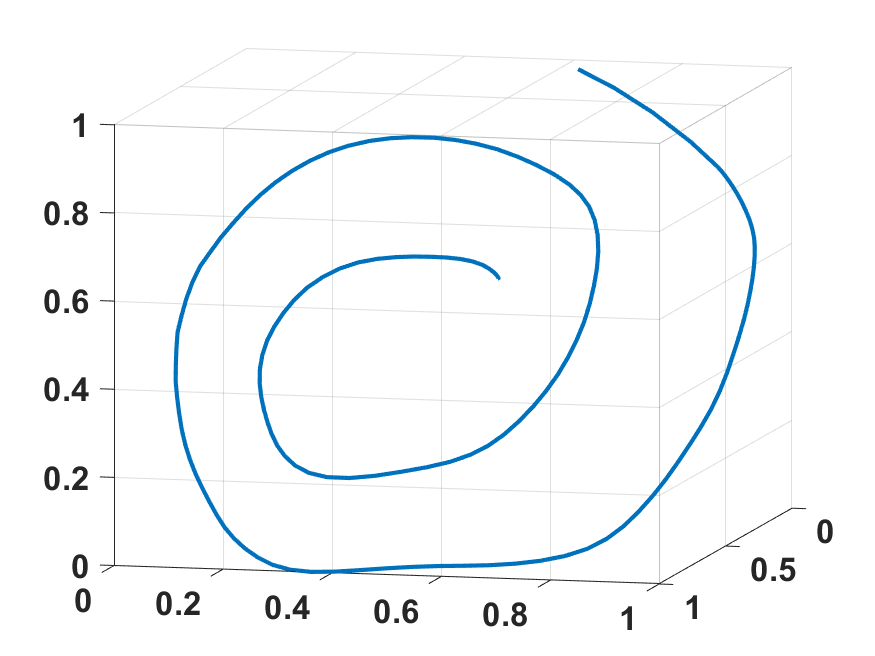}}
\subfloat[``Triangle'']{\includegraphics[width=0.22\textwidth]{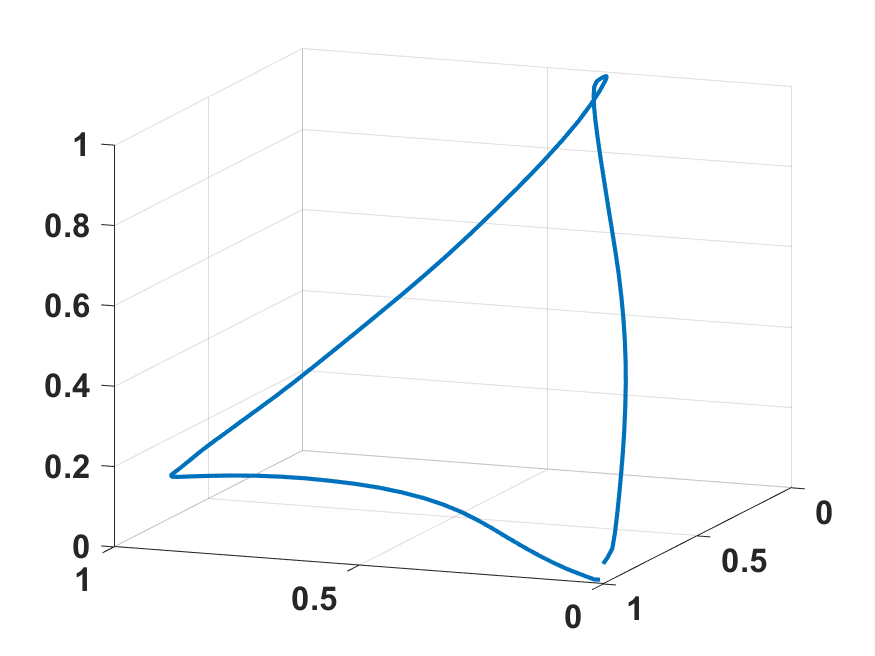}}

\caption{Designed tasks with curve lines, straight lines, and sharp curve characteristics. }
\label{fig:task graphs}

\end{figure*}

\section{Experiments}
In this section, we describe the experiments conducted on the motion-controlled robotic avatar platform to evaluate our emotion classification approach, including emotion stimulation, motion selection, and experimental setup.

\subsection{Emotion Stimulation}
\label{Emotion stimulation}
Emotion stimulation is an established method for obtaining high-quality emotion data~\cite{10.1145/2973750.2973762}. Generally, emotional responses to music and emotion thresholds vary significantly from person to person~\cite{kim2008emotion}. Personal experiences and cultural background influence people's reactions to the same music~\cite{kim2008emotion,10.1145/2973750.2973762}.
Thus, we followed an established personalisation approach: participants were allowed audio files of their choice, such as songs, noise, or even stand-up comedy recordings to stimulate each corresponding emotion individually~\cite{10.1145/2973750.2973762, kim2008emotion}.
While choosing not to strictly control the emotional stimuli could add variance, we chose the personalised approach in the absence of any truly consistent way to elicit emotion in people, a limitation which impacts the entire field.

\subsection{Non-Stylized Motions}

Our work focuses on the functional, non-stylized motions that are likely to be performed during telerobot control, rather than motions intended to express emotion~\cite{loghmani2017emotional}. 
Specifically, we designed 14 non-stylized motions as tasks and divided them into two categories. The first category was mid-air gestures, which are performed in many motion-controlled robotic avatar scenarios to perform matching motions in the 3D space. 
The second category is the line-tracing tasks, commonly performed in motion-critical scenarios. Operators are required to move the robotic avatar to follow pre-designed trajectories, which restrict their motions.  

\subsubsection{Mid-air Gestures}
We designed nine mid-air gestures: (1) cursive ``Lw'', (2) ``Star'', (3) ``Stir'', (4) ``S'', (5) ``Triangle'', (6) ``Drinking'', (7) ``Knocking'', (8) ``Throwing'', and (9) ``Waving''. The mid-air gestures (6) to (9) were chosen as examples of social HRI and freestyle tasks in remote education scenarios \cite{granados2017dance, casan2015ros}. These were common daily-life gestures that could occur in everyday situations \cite{7989198}, encompassing scenarios like sharing social activities (``Drinking'', `Throwing'') or interpersonal social interaction (``Knocking'', `Waving''). The participants controlled the robotic avatar to perform these mid-air gestures in a non-prescriptive manner without hard constraints.

\subsubsection{Line-tracing Tasks}
We also designed five line-tracing tasks which contain typical features of motions that would appear in motion control scenarios including (1) ``Lw'', (2) ``Star'', (3) ``Stir'', (4) ``S'', and (5) ``Triangle'' (see Figure~\ref{fig:task graphs}). 
For example, ``Star'' and ``Triangle'' contain sharp turning points. ``S'' contain smooth turns, while ``Stir'' contains consecutive turns.
Cursive ``Lw'' synthesizes all the features of the other drawing tasks.

Compared with the mid-air gestures, the line-tracing tasks required the participants to follow the printed trajectory reference, such as robotic-assisted spine surgery~\cite{d2019robotic}, total knee arthroplasty ~\cite{hampp2019robotic} and dental implantology~\cite{wu2019robotics}. To be more specific, we add constraints to these tasks, which simulate mission-critical control scenarios.
\par

In this work, we followed an approach of Huang et al.: utilising a classification of foundational tasks that can be widely applicable to different telerobotic scenarios~\cite{huang2021robot}. 
We included a selection of task types representative of the core movements of real-world tasks, including surgery, education, and social scenarios, as featured in prior work~\cite{d2019robotic, casan2015ros} to provide a good foundation. However, our study has a necessarily narrower scope than real-world use, which should be addressed in the future by co-designing tasks with real-world telerobotic users, to ensure specific and highly ecologically valid tasks for specific domains.

\subsection{Experimental Setup and Data Collection}
\label{subsec:experimental_setup}

We recruited ten volunteers (3 female, 7 male) to conduct the experiments. All were university students with ages ranging from 20 to 25 (mean = 24.3, $\sigma$ = 1.42). The volunteers first read an information sheet, signed a consent form, and brought their individual pre-pickup audio clips (see Section~\ref{Emotion stimulation}) to trigger emotional responses (joy, pleasure, annoyance, and sadness). Each volunteer first underwent 30 minutes of training to become familiar with the control process.
Then, participants took a 10-minute break to help them reset to a more neutral emotional state.
The experiments were conducted using the motion-controlled Universal Robot UR3e platform, as introduced in Section~\ref{sec:platform}, deployed in a quiet $50 m^2$ laboratory room. Six OptiTrack cameras are placed in a $5m$-by-$5m$ square area. Participants wore a marker glove on their right hand, and ECG devices were attached to their other hand and the ipsilateral ankle, in order to minimize movement. During the experiments, participants sat in the centre of the OptiTrack cameras to control the UR3e while listening to the audio they selected. Operators were given three minutes of emotional stimulation at the beginning of each task, and the collection time was approximately 2.5 minutes per task. A 10-minute emotional recovery break was given between each task. The order of emotion stimulation was random to offset the influence between different emotions. At the end of each task, an interview was conducted to assess if the stimulated emotions were consistent with the target emotions. If so, the collected data were labelled with the subject’s reported emotion.
As ``Lw'' synthesized all the drawing motions tasks, we required all participants to perform both the line-tracing ``Lw'' and in-air ``Lw'' under all five emotions (joy, pleasure, sadness, annoyance, and neutral). Each task was performed 15 times under each emotion. Given this, there were 1500 total instances of the ``Lw'' task (10 subjects $\times$ 5 emotions $\times$ 15 times $\times$ 2 non-stylised tasks types). 

Then, five of the ten participants are asked to perform the remaining twelve tasks, which include eight mid-air gestures (``Star'', ``Stir'', ``S'', ``Triangle'', ``Drinking'', ``Knocking'', ``Throwing'' and ``Waving'') and four line-tracing tasks (``Star'', ``Stir'', ``S'' and ``Triangle''). These five participants performed each task 15 times under each emotion, resulting in 4500 (5 subjects $\times$ 5 emotions $\times$ 15 times $\times$ 12 non-stylised task types) task instances. 
We did not ask the remaining 5 participants to complete these extra tasks due to time constraints per experimental session, as these participants took significantly longer to complete the ``Lw'' task. In total, 6000 task instances were collected from the experiments to serve as the dataset.




\subsubsection{ECG data collection}
As mentioned in Section~\ref{subsubsec:physiological_signals}, ECG is a well-established emotion recognition method, so we collected ECG data to compare its suitability to our method.
There are reasons why ECG may not be ideal for telerobotic scenarios: ECG signal collection for emotion classification is normally taken for approximately 8 minutes ~\cite{10.1145/2973750.2973762,kim2008emotion} and can be confounded by motion artefacts. 
In telerobotic motion control scenarios, the subject will often move, and tasks can be quite short, such as pick and place or stirring tasks~\cite{rakita2020effects,mainprice2015predicting,perez2015fast}. 
Even in longer tasks, such as teledriving, the inherent motion of the task has led to ECG classification being limited in past work~\cite{el2019random, akbas2011evaluation, keshan2015machine, munla2015driver}.
However, as discussed in the Related Work, there is no prior work to show how viable ECG could be in telerobotic scenarios and how much influence the motion artifacts will have.
We sought to understand whether our proposed approach could prove a more suitable alternative, and so compared the two in this work.
To assess this, we outfitted participants with an integrative ECG~\cite{Thornburg05} Attys biomedical sensor device on the hand they did not use for operation and instructed them to keep this hand as steady as possible, as well as the ipsilateral ankle to minimise movement artefacts.
The ECG signals collected during task execution were then used to implement traditional ECG-based emotion classification, to compare its suitability with our proposed robotic avatar emotion classification approach.

\begin{figure*}[t]
\centering
\subfloat[a][Trajectory and joint data
comparison]{\includegraphics[width=0.28\textwidth]{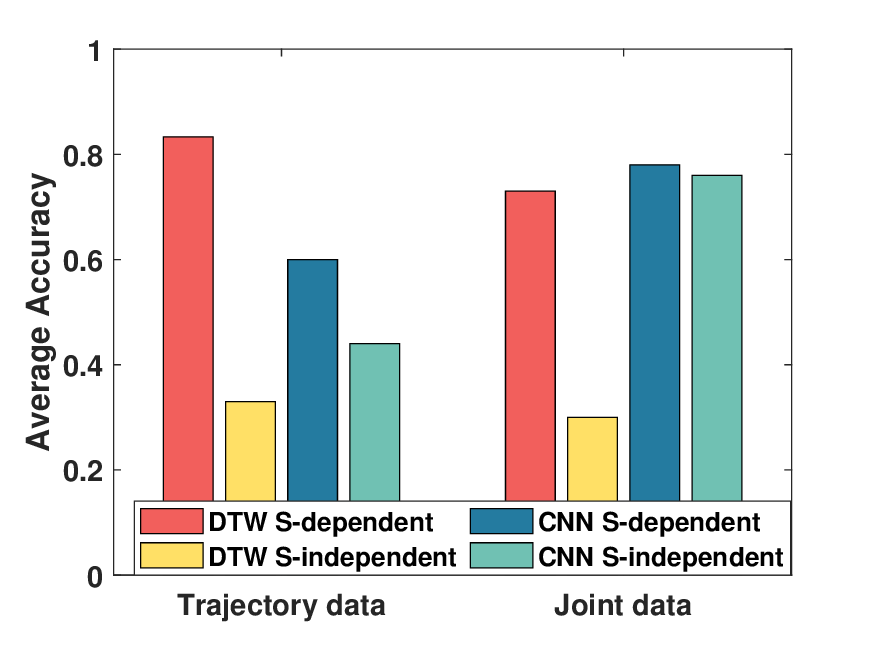}\label{fig: t_j}}
\subfloat[b][DTW and CNN algorithm
comparison]{\includegraphics[width=0.28\textwidth]{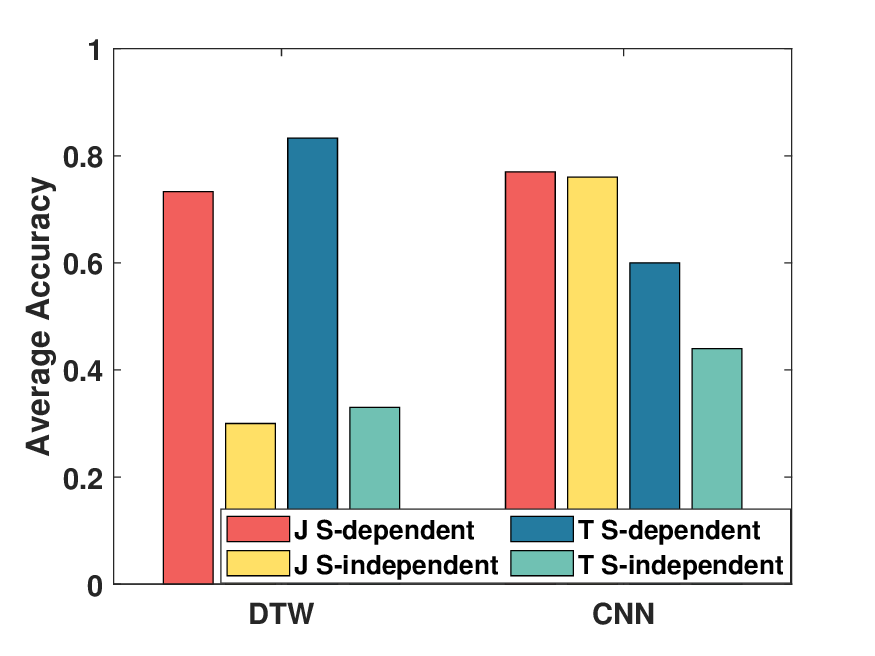}\label{fig: D_C}}
\subfloat[c][Subject dependent and in-
dependent model comparison]{\includegraphics[width=0.28\textwidth]{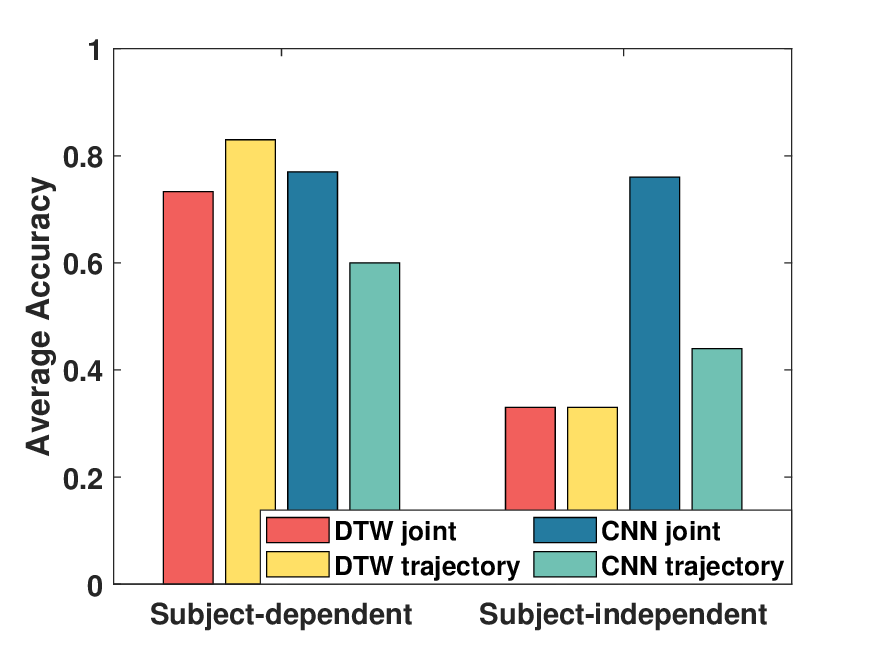}
	\label{fig: d_i}}
\caption{The emotion classification results for different classifiers trained by different algorithms and different data (``S'' stands for ``Subject'', ``J'' stands for ``Robot Joint Data'', and ``T'' stands for ``Robot Trajectory Data'').}
\label{fig:three graphs}    
\end{figure*}

\section{Results and Analysis}
\label{sec:results}

In total, we collected 6000 instances to evaluate our emotion classification approach, in-line with or exceeding prior work~\cite{10.1145/2973750.2973762, loghmani2017emotional}.
The dataset can be accessed on the online repository Figshare\footnote{Figshare data repository: doi.org/10.6084/m9.figshare.28281086}.


Following task classification~\cite{huang2021robot}, we trained two emotion classifiers for each task category: a subject-dependent classifier trained and tested on the dataset of each individual subject, and a subject-independent classifier trained using a leave-one-subject-out procedure.
Subject-dependent classifiers study each person's individualized emotional expressions, better accounting for individual emotion thresholds and allowing for more accurate detection via personalisation~\cite{wu2022estimating}. 
Subject-independent classifiers have applicability to use cases in which classifiers are pre-trained on the common emotion information of operators. 
We used a DTW-based algorithm (Sec.~\ref{subsec:dtw}) and a CNN-based algorithm (Sec.\ref{subsec:cnn}), respectively, to train the subject-dependent and subject-independent classifiers. 
The classification results per classifier, algorithms, and data are shown in Figure~\ref{fig:three graphs}
\par
On average, the CNN-based algorithm performed best (Fig.~\ref{fig: D_C}), as the DTW-based algorithm showed worse performance when training the subject-independent classifier.
The DTW-based algorithm achieved greater performance using the robot's end-effector trajectory data, while the CNN-based algorithm performed better with joint data (Fig.~\ref{fig: t_j}). 
The accuracy of subject-independent classification was lower than subject-dependent classification, as expected (Figure~\ref{fig: d_i}). 
Based on these results, we used the DTW-based algorithm and robot end-effector's trajectory data to train the subject-dependent classifier and used the CNN-based algorithm and robot joint data to train the subject-independent classifier. 
Individual differences in emotive motion may explain the lower performance of the DTW algorithm for subject-independent data. 
For example, one person may move faster or further than another, even when both are annoyed. 
The DTW algorithm measures the distance difference between two different emotional instances, so different expressions of emotions between people make the emotion instances less comparable, and finding common emotion information across people more difficult.
The emotion classification performance of these two types of classifiers is presented below.

\begin{table*}[b]
\caption{Subject-dependent emotion classification for each of the mid-air gestures and line-tracing tasks.
}
\label{tab:task}
\begin{center}
	
	\begin{tabular}{|c|c|c|c|c|c|c|c|c|c|}
		\hline
		Tasks & Lw& Star&Stir& S& Triangle&Drink &Knock&Throw&Wave\\
		\hline
		Mid-air gestures&0.851&0.853&0.817&0.913&0.920&0.807&0.860&0.858&0.909\\
		\hline
		Line-tracing tasks&0.777&0.737&0.757&0.833&0.791&NA&NA&NA&NA\\
		\hline
	\end{tabular}
	
\end{center}
\end{table*}

\subsection{Subject-Dependent Results}

\subsubsection{Classification Performance Variance By Subject}
\begin{figure}[t]
\centering
\includegraphics[width=0.45\textwidth]{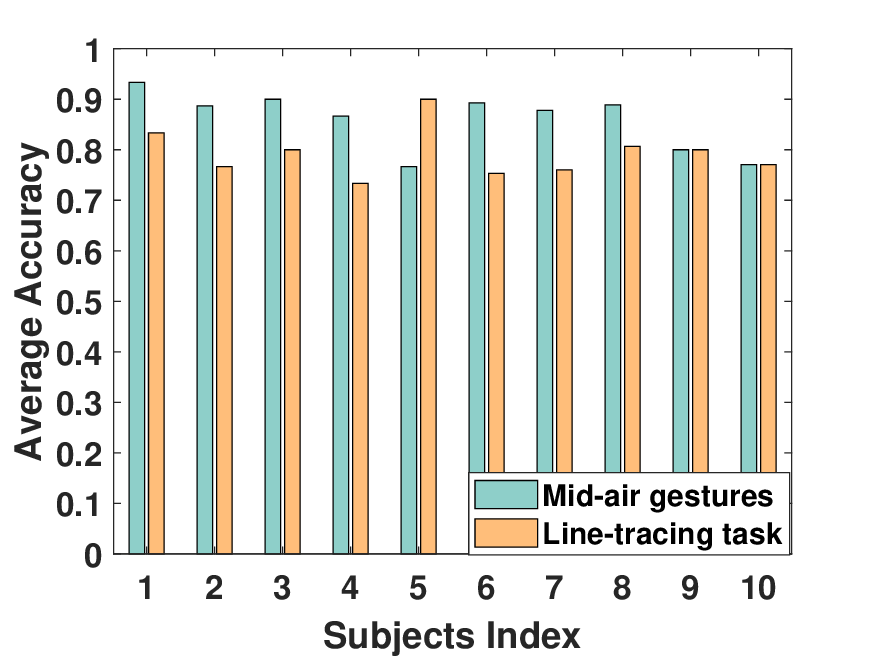}
\caption{Subject-dependent emotion classification of mid-air gestures and line-tracing tasks for ten subjects.}
\label{fig:acc}

\end{figure}

Figure~\ref{fig:acc} shows the average emotion classification result of all mid-air gestures and line-tracing tasks, for each subject respectively. Overall, the average accuracy of subject-dependent classification among all subjects was 83.3$\%$. We can observe that the performance on mid-air gestures and the line-tracing task is similar among different subjects, which indicates the proposed approach works for different types of motions. The results show our approach can classify each operator's emotions with relatively high accuracy.

\subsubsection{Classification Performance Variance By Task}
The subject-dependent classifier's average performance among all the subjects for each mid-air gesture and line-tracing task is presented in Table~\ref{tab:task}. The average accuracy achieved by the mid-air gestures was 86.5$\%$, while the average accuracy of the line-tracing tasks was 77.9$\%$. Performance of the ten drawing tasks ranged from 73.7$\%$ to 92.0$\%$, with the ``S'' task showing the best performance in both the mid-air gestures and line-tracing tasks. Performance of the four social tasks ranged from 80.7$\%$ to 90.9$\%$, with the ``Wave'' task performing best. The results show that our subject-dependent algorithm can be used to infer emotions from a wide range of tasks performed by different users.

\subsubsection{Classification Performance Variance By Emotions}
The average detection accuracy for each type of emotion across tasks performed by each subject is presented in Table~\ref{tab:ed}. Our approach detected all five types of emotions with an average accuracy of 83.3\%. In particular, the average detection rate for ``Joy'', ``Sadness'', ``Annoyance'', ``Pleasure'' and ``Neutral'' among the ten subjects is 83.12\%, 86.67\%, 90.75\%, 68.11\% and 87.31\%, respectively. The results show that this approach generally works for detecting different types of emotions. ``Pleasure'' was something of an outlier with worse performance, which may indicate that being in this high valence, low arousal affective state resulted in less distinct and expressive movement features than the other emotional states, particularly Joy (high valence \& arousal).

\begin{table}[b]
\caption{Subject-dependent classifier's average emotion detection accuracy for different subjects. }
\label{tab:ed}
\begin{tabular}{|c|c|c|c|c|c|}
	\hline
	\makecell{Subject \\No.}  &Annoyance&Pleasure& Sadness& Joy&Neutral\\  
	\hline
	
	1&0.9167  &  0.7500  &  0.9167  &  0.9167 &   0.9167  \\\hline
	2&0.8488  &  0.8750  &  0.8310  &  0.7952 &   0.8690   \\\hline
	3&1.0000  &  0.5500  &  0.9000  &  0.9000 &   0.9000   \\\hline
	4&1.0000  &  0.3333  &  0.9167  &  0.9167 &   0.8333   \\\hline
	5&0.9167  &  0.8333  &  0.7500  &  0.6667 &   1.0000   \\\hline
	6&0.8810  &  0.7381  &  0.8810  &  0.8571 &   0.8571   \\\hline
	7&0.9286  &  0.7619  &  0.8452  &  0.7024 &   0.9405   \\\hline
	8&0.8214  &  0.7452  &  0.9262  &  0.8762 &   0.9286   \\\hline
	9&0.9167  &  0.5000  &  0.9167  &  0.9167 &   0.7500   \\\hline
	10&0.8452 &   0.7238 &   0.7833 &   0.7643&    0.7357  \\\hline
	Mean&0.9075 &  0.6811  &  0.8667 &   0.8312  &  0.8731\\\hline
	SD &0.0608 &0.1674 & 0.0621 &0.0936 &0.0832 \\
	\hline
\end{tabular}

\end{table}

\subsubsection{Classification Performance Variance By Number of Emotions}

\begin{figure}
\centering

\subfloat[Subject-dependent classifier.\label{fig:emotion types 1}]{\includegraphics[width=0.33\textwidth]{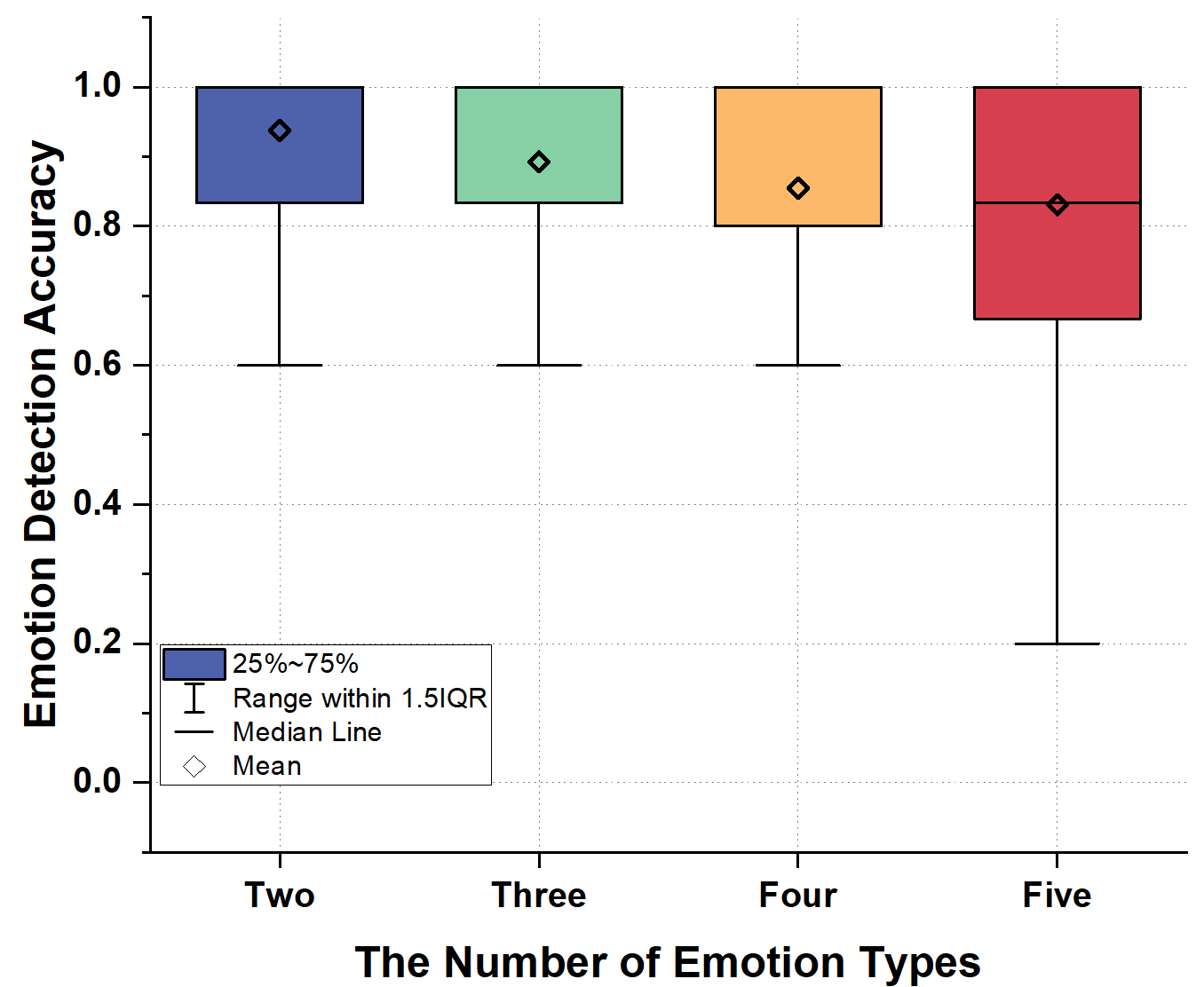}}
\\
\subfloat[Subject-independent classifier.\label{fig:emotion types 2}]{     \includegraphics[width=0.33\textwidth]{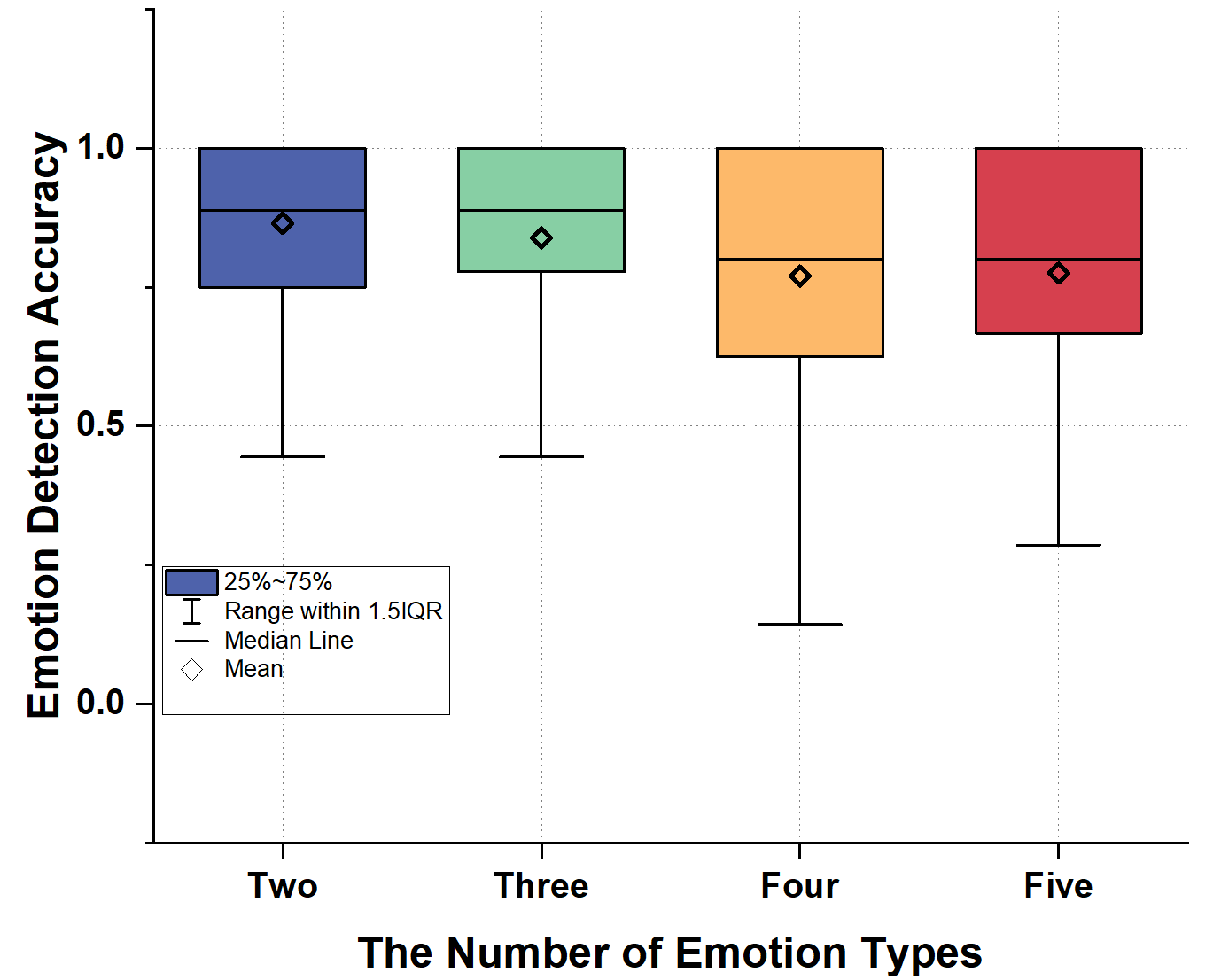}}
\caption{Emotion classification results regarding different numbers of emotion types for different classifiers.}
\label{fig:emotion types}

\end{figure}

Figure~\ref{fig:emotion types 1} shows the average emotion classification results among all tasks and subjects when different numbers of emotion types are involved. When classifying two types of emotions (\textit{e.g.}annoyance and neutral), our approach achieved an average accuracy of 94.05$\%$. When classifying three types of emotions (\textit{e.g.}joy, annoyance, and neutral), the average performance degraded to 89.8$\%$. And when classifying four types of emotions across all types, the performance further degraded to 86.3$\%$. When there are five emotions (joy, pleasure, sad, annoyance, and neutral) to be classified, our approach can still achieve an accuracy of 83.3$\%$. The results demonstrate that our approach is able to classify common emotions with relatively high accuracy.

\subsection{Subject-Independent Results}
Trajectories of tasks performed by different operators under the same emotion showed different motion features, thus, it is hard to use the robot end-effector's trajectory to implement subject-independent emotion classification. Instead, we use the robot joint data to train the subject-independent classifier introduced above.

Intrinsically, subject-independent emotion recognition is more of a challenge as the way subjects express emotions varies, which can be seen manifesting in the performance variance of the DTW  (see Section~\ref{sec:results}). As expected, subject-independent performance was lower than that of subject-dependent. 
To study the capability of our system to recognize the emotions of users not included in the training model, we utilised a leave-one-subject-out cross-validation procedure (each subject was left out of the training dataset for their testing), increasing the generalisability of results and reducing the risk of over-fitting.
Across tasks and affective states, this analysis achieved an average emotion recognition accuracy of 76.5$\%$. The following sections discuss accuracy variance between tasks and affective states.

\subsubsection{Classification Performance Variance By Task}

The subject-independent emotion classification for each mid-air gesture and line-tracing task is shown in Table~\ref{tab:task-ind}. The ``LW'' task was trained and tested on 10 subjects while the rest were trained and tested on 5 subjects.
Our approach had an average accuracy of 76.8$\%$ and 75.8$\%$ for the mid-air gestures and line-tracing tasks, respectively. The emotion classification performance ranges from 66.4$\%$ to 85.2$\%$, which is comparable to the results of the existing work~\cite{loghmani2017emotional} that used human motion signals for emotion classification. 

\begin{table*}[t]
\caption{Subject-independent emotion classification for each of the mid-air gestures and line-tracing tasks.}
\label{tab:task-ind}
\begin{center}
	\begin{tabular}{|c|c|c|c|c|c|c|c|c|c|}
		\hline
		Tasks & Lw& Star&Stir& S& Triangle&Drink &Knock&Throw&Wave\\
		\hline
		Mid-air gestures &0.775&0.734&0.757&0.758& 0.828&0.711& 0.789&0.719&0.844 \\
		Line-tracing tasks&0.682&0.828&0.664&0.766&0.852&NA&NA&NA&NA\\
		\hline
	\end{tabular}
\end{center}
\end{table*}

\subsubsection{Emotion Detection Accuracy}
As shown in Table~\ref{tab:ed-ind}, our subject-independent method achieved 76.5$\%$ emotion detection accuracy across the five emotions. Interestingly, our approach achieved higher accuracy for emotions with lower arousal. A possible reason for this could be that low arousal emotions may contain more subject-independent features than high arousal states. 

\begin{table}[b]
\caption{Average emotion detection accuracy achieved by the subject-independent method.}
\label{tab:ed-ind}
\begin{tabular}{|c|c|c|c|c|c|}
	\hline
	Emotion& Annoyance &Joy& Sad&Pleasure&Neutral\\  
	\hline
Accuracy&0.676&0.779&0.870&0.700&0.779\\
\hline
\end{tabular}

\end{table}

\subsubsection{Classification Performance Variance By Number of Emotions}
Figure~\ref{fig:emotion types 2} shows the performance of our subject-independent method when there are different numbers of emotion types. Specifically, the average emotion classification accuracy is 88.3$\%$ for two emotions (\textit{e.g.}, annoyance and neutral), 84.0$\%$ for three emotions (\textit{e.g.}joy, annoyance, and neutral), 80.9$\%$ for four emotions (\textit{e.g.} joy, sad, annoyance, and neutral), and 76.5$\%$ for all five emotions (joy, pleasure, sad, annoyance, and neutral).

\subsection{Our approach versus ECG-based emotion recognition}

As we discussed in Section~\ref{subsubsec:physiological_signals}, ECG-based emotion classification uses minor changes in physiological signals (heart rate) to detect emotional changes and requires physical contact sensors to observe the ECG signal. These two factors constrain the application scenarios of the ECG-based emotion classification. 
In order to compare the suitability of our emotion recognition method, we collected the ECG data of the subjects during the study. The ECG dataset size is the same as the robot motion, so the results of ECG and the robot are comparable.  The time length of each ECG instance was around 210s. We used IIR to filter the ECG data and extracted the emotion-related features as mentioned in ~\ref{subsec:ECG ER}. As expected, the average accuracy across ten subjects was 56.6$\%$ lower than the existing results 87$\%$~\cite{10.1145/2973750.2973762}, likely due to both motion artefacts and measurement duration.

\section{Discussion}

Our work shows that a robotic avatar's motion can be used to infer the operator's emotions. 
We know now that emotions can have a vital role in the interaction between humans and robots, as they have a direct impact on the control of the remote robot. Thus, it is beneficial to understand and observe operator emotions to avert erroneous operations in safety-critical scenarios.

\subsection{Current Performance, Limitations and Next Steps}

\subsubsection{Current Performance of our Approach}
\qquad  
\par
\noindent\textbf{Emotional Information Involved in Interaction}

Our robotic avatar can inherit human hand trajectories but cannot reproduce human trajectories perfectly. On the one hand, the skeleton and the degree of freedom (DOF) of the robot arm and the human arm are different. On the other hand, the limitation of the control algorithm and the communication delay cause a deviation between the robot's and the human's trajectory. Although the robotic arm can only reproduce lossy trajectories, our emotion classification (dataset = 6000, n = 5, accuracy = 83.3$\%$) outperforms prior work using individual human status data (dataset = 235, n = 5, accuracy = 70.05$\%$)~\cite{loghmani2017emotional}. 
This indicates that measuring emotion using our methodology via inference of robotics arm trajectories may be as, or more sensitive than prior approaches.
Both our work and Loghmani et al.~\cite{loghmani2017emotional} used non-stylized motions, but our participants performed in an interactive control scenario, i.e., operators observe a robot's movements to adjust their own behaviours in real-time. It could be that emotional expressions are more pronounced in such interactive control scenarios. The influence of interaction scenarios on emotion expression should be further explored in the future, which could profoundly impact subsequent interaction design. 

\par
\noindent\textbf{Suitability of Telerobot Emotion Classification Compared to Traditional Methods}

We attached the ECG device to the operators' stable hands and ankles to capture their heart rate signals. We found this to be a limitation of the ECG experiment setup, which requires longer measurement time and for operators to remain stable for optimal performance, as in motion-controlled scenarios, users need to move and are unlikely to remain still enough. Similarly, many existing methods of using physiological signals to classify emotions require humans to stay stable, which limits the application of ECG emotion classification. 
While applicable in controlled laboratory experiments, these limitations would preclude the real-world use of these techniques in telerobot scenarios.
Our work verifies, for the first time, the limitation of ECG in human remote-control robot scenarios and shows that this method lacks ecological validity for this use case. 
For example, when we evaluate whether a driver is exhausted or not, we can not require him/her to stay stable while driving. 
This motivates an alternate approach, such as the approach we explored in this work, utilizing telerobotic motion to infer emotions. From this point of view, behaviour-based emotion classification is more practical and promotes many applications in this field.  

\subsubsection{Limitations and Future Work}
In this study, we used audio files, a well-established emotion stimulation method~\cite{marin2018affective}, to elicit emotions and interviewed participants to check they felt the correct emotion was evoked. 
This approach has limitations, however, as it has been regarded as non-immersive when immersion is an important aspect of eliciting emotions in real experiences~\cite{marin2018affective}. 
Future work could seek to adopt a more immersive emotion elicitation approach, such as leveraging Virtual Reality~\cite{marin2018affective}.

Another core limitation of this work is that emotion is inherently ambiguous and complex, so there may exist disagreements between participant annotators' labels and their real emotions ~\cite{wu2022estimating}. 
In addition, while we took steps to help participants regain a neutral emotional state between tasks, this cannot be fully controlled.
While we used established methodology in this work, this is a general problem within the field of affective computing ~\cite{hsu2017automatic}.
\par
Our work features a participant sample size of 10, with some tasks only performed by 5, which limits the immediate generalisability of our current model to the wider population and real-world applications.
We did, however, collect 6000 instances in total, a larger set than similar prior affective computing studies~\cite{10.1145/2973750.2973762,loghmani2017emotional}, and we achieved comparable results. This demonstrates the feasibility of our emotion inference method. 
In the future, however, a larger participant pool and a wider set of tasks are required to fully understand the efficacy of our approach in realistic remote-operation scenarios. 
Our initial promising findings motivate this further exploration, but much work is still required to develop and assess a truly ecologically valid implementation of our approach.
Emotional expression may also vary in intensity in different tasks. For example, linear motions may show fewer emotive features than complex motions. This should be accounted for when aiming to achieve real-world generalisability. Similarly, future work could adopt differentiation between tasks used in training and testing to further explore applicability to unseen real-world tasks.
Participants also received 30 minutes of training, which is necessarily limited compared to a full training regime for real-world telerobotic operation.
\par
In this work, we used an ECG device, but found it unsuitable for telerobotic scenarios, as ECG data collection suffered from motion artifacts in the data. This intrinsic limitation made it difficult to capture effective emotional information, and thus, it was difficult to compare its efficacy with our method. Future work should seek to utilise additional sensors for multimodal data, such as EEG and respiration rate, which could achieve accuracy comparable with that previously shown by ECG~\cite{hasnul2021electrocardiogram} while minimising the impact of motion artefacts~\cite{abdullah2021multimodal}.
This would allow for a more robust comparison with this novel robot motion-based inference approach.
Furthermore, direct comparison with user-side motion-capture models would also better inform the effectiveness of our approach~\cite{BeyanMoCap}.

\subsection{Implications for Current and Future Telerobotic Applications}

\subsubsection{Applied Remote Robot Operation:}

Our work found that operator emotion can be inferred from telerobotic movement and that certain emotions can result in more vigorous and pronounced movement.
Classifying based on the robot arm movements carries the core advantage of allowing one to observe and calibrate the effects of emotive user input by observing the ground truth of the consequences of these emotions, which is the robot's motion.
It is prudent to consider how this might impact current telerobot applications differently.
For example, telerobotic keyhole surgery is an extremely precise and safety-critical environment where small movements could have dire health consequences. 
Given this, intervening swiftly to remove control during moments of heightened operator emotion could be highly beneficial. 
This would render the end-effector suddenly stationary, which is unlikely to be consistently dangerous, as keyhole surgery is made up of prolonged pauses and slow movements, but could be problematic if the effector is currently interacting with tissue. 
Furthermore, keyhole surgeons are highly specialised, so handing over control to a replacement operator may be difficult.
\par
By contrast, teledriving presents a more difficult scenario for intervention following emotion detection.
While prior work has observed driver emotion directly from onboard control telemetry~\cite{zepf2020driver}, it is still unclear how this information should be applied to reduce danger.
While driving is a less precise task than telesurgery, it is still safety-critical, and erratic behaviour may warrant the removal of control. Unlike telesurgery, however, removing control of teledriving leaves a vehicle that is still in motion. 
Given this, an intervention may need to either hand over to an autonomous driving system or perhaps reduce noise in the operator's control, rather than remove it (see \ref{AI-dampening}).
Another context to consider is industrial applications, such as the telerobotic handling of nuclear waste containers. While also safety-critical, this operation is less precise than surgery, making the emotional level required to intervene more extreme, and control may be safely paused and handed over to another remote technician to complete the task.
Beyond the differing practical concerns of observing operator emotions and intervening in different contexts, we must also consider the potential impact on these humans in the loop and how the system can be designed to be supportive, rather than combative.
\par
~More enhanced feature extraction and advanced machine learning techniques are required in the future to generalise tasks across varied operations in the real world. Inherited affective movements may vary depending on the input device and the robot system used. Outside of using motion tracking to control remote robots, alternative control schemes, such as haptic gloves and physical controllers, have been applied to control remote robots. It would be valuable to explore how our method of emotion inference performs across these input devices. Transfer learning-based methods could be developed~\cite{torrey2010transfer, lin2017improving, lu2023hybrid} that can apply previously learned knowledge from available large-scale data and establish a new model. Such approaches could be used to achieve cross-subject emotion recognition.
Finally, in this work we explored a set of four basic emotions from each quadrant of the circumplex model as a foundation. To best suit the needs of specific applications, future work should explore training models to focus on application-specific emotional states, such as confusion and frustration, in consultation with field experts.

\subsubsection{Understanding the Human Impact:}

While emotional inference could be used to intervene during safety-critical telerobotic scenarios, possible pitfalls must be considered. 
First is the issue of privacy. 
Safety-critical telerobotic operation supports various applications, including those for which humans cannot be physically present, such as nuclear waste handling, as well as healthcare, transport, and industry.
Working remotely can afford employees increased privacy when compared to those to work on-site, which they may value~\cite{Neustaedter2006,Wutschert2021}. 
Operators may feel that having their emotional state inferred through robotic arm movement infringes on their right to work while managing their private internal emotional state.
While co-located workers would naturally display emotional cues through their body language or voice, systematically monitoring and using their emotions to assess performance or intervene for safety reasons would require fitting with traditional electric-signal-based monitoring devices.
As discussed in Section \ref{background:emoreg}, such devices can be confounded by the movement inherent in telerobotic operation. Our system could, therefore, offer a functional replacement for this context. 
While it would be clear to an employee that they have been fitted with a wearable monitor, it may be less clear that they are being monitored based on robotic avatar movement. Thus, this system should be clearly signposted and the informed consent of operators obtained.
\par
Another issue operators could experience is fear of loss of agency, as they know their control could be removed due to automatic inference of their involuntary emotional state, which could in turn lead to an adversarial relationship between user and system.
For example, operators may seek to practice emotion regulation using real-time Response Modulation \cite{Gross2002} in order to avoid losing control, which in turn may deplete attentional resources and risk worsening performance. Losing control, when they otherwise would not have, could also damage an operator's confidence and mental well-being. If the loss of control is observable by peers, it may also lead to perceptions of incompetence or feelings of shame.
\par
Repeated interventions or interruptions by such a system could also be seen as frustrating or annoying.
Additionally, emotional thresholds and the impact of heightened emotions on performance will also differ between individuals, potentially reducing the system's generalised recognition accuracy and leading to false positives.
False positive identification of risky emotional states may cause inefficiencies or frustration. For example, in real-life telesurgery, an intervention resulting from false-positive identification of an individual's emotions may break the surgeon's rhythm and lead to negative outcomes.
This further motivates the operator-personalised training approach we explored in this work (see Sec 6.1).
Another potential advantage of a ground-truth observation approach is that if a risky emotional state is detected it can be compared to the resultant movement to double-check if there is a true potential for harm.
\par
Either way such a system should be implemented in an ethical and empathetic manner, with the removal of control used as a safety-driven last resort, in order to mitigate users harbouring resentment for the system.
As an aside, there are less disruptive ways emotional information during interaction could be used, such as evaluating the operator satisfaction as feedback to improve the robot's control algorithms.
In the next section, we propose an alternative moderate approach to removing operator control, emotive-motion dampening, which could mitigate these issues while still improving safety outcomes.

\subsubsection{Future Applications:}
\qquad  
\par
\noindent\textbf{AI-Assisted Emotive-Motion Dampening}
\label{AI-dampening}

As discussed, emotional influence on telerobotic avatar movement could have negative safety outcomes, but simultaneously the sudden removal of operator control based on their emotional state could have negative practical and psychological ramifications.
Given this, we propose an intermediate solution, the real-time dampening of emotive-motion features.
When enabled, real-time AI would be leveraged to filter out the drastic and jerky features of user input motion that are caused by a high-arousal emotion state, normalising to a smoother trajectory. 
This approach is analogous to the aim-assist feature used in some first-person video games~\cite{10.1145/2556288.2557308}, or with \textit{shared control} paradigm explored in prior work~\cite{gopinath2016human, javdani2015shared, reddy2018shared}, whereby control is shared between the human and the robotics autonomy.
Extending prior work, we propose to apply this technique responsively based on the operator's inferred emotional state. 
\par
In some scenarios, this could be enabled by default, although in high-precision scenarios, such as telesurgery, it could reduce the operator's level of fine-grain control.
In these scenarios, the dampening system could instead be enabled only when an emotional state that could compromise the safety-critical task is detected.
Such a system could also have privacy benefits, as normalising robotic avatar motion could be used to prevent further observation of the operator's emotions. 
The calibration of such a system and its impact on different telerobotic tasks would be valuable topics for future research.

\noindent\textbf{Emotional Intelligence Encounters with Robotic Avatars}

By using similar emotion inference approaches, we could facilitate the recognition of naturalistic body language, trained on real human motion data, in both virtual and physical robotic avatars. 
VR allows people to be embodied within virtual environments and act within them using virtual avatars which can express their body language. Liebers \textit{et al.}~\cite{liebers2021understanding} found it is possible to identify individuals via their virtual body language in VR and it has been shown that virtual agents can express emotion through body language ~\cite{nayak2005emotional, pelachaud2009modelling}. Our methods could be applied to these virtual avatars, allowing for the automatic detection of users' emotions in VR settings. 
This could be used to tailor user experiences; for example, if during a VR game a user is expressing anger or sadness the game could dynamically become calmer~\cite{Yun2009}.
Furthermore, our approach could be leveraged to enable more emotionally intelligent interactions with the virtual world, NPCs and other users.  
\par
These emotionally intelligent encounters could also take place in real-world settings.
In the wake of the COVID-19 pandemic, working from home has become more prominent as a current and future labour trend. In the future, we may see physical robotic avatars, such as robotic arms, partially or fully replace human workplaces such as offices or factories.
If these avatars could both express emotion and have their operator's emotion understood by other co-located humans or robotic avatars, it would help maintain affective relationships commonplace in social and working contexts.
Our work shows that robot arm avatars have inherently distinct movement traits from differing operator emotions.
Humans already possess the ability to infer affect from human arm movement~\cite{pollick2001perceiving} and future work could now explore if this also extends to the emotive movements of robotic arms inherited from their operators. 

Finally, future work could investigate how other robotic form factors may inherit emotions, such as quadruped robots\footnote{https://bostondynamics.com/products/spot/ }, limited humanoid social robots such as \textit{Pepper}\footnote{ https://www.aldebaran.com/en/pepper} and \textit{Sophia}\footnote{https://www.hansonrobotics.com/sophia/  } which can only articulate their heads, arms and torso, or robotic hands, such as \textit{Shadow Hand}\footnote{ https://www.shadowrobot.com/ }. 
While telerobotic avatars inherently express emotion from the operator's natural arm movements, these quadrupedal or social robots are instead operated using a controller, such as a gamepad, so whether emotion can be inferred from such control mechanisms should also be investigated.
Finally, future full-bodied robotic avatars could inherit yet more complicated and nuanced emotional features, as more pronounced movement across the whole body is used to express emotive movement features and the relationship between different body parts can provide more emotional information, as has been explored with virtual agents~\cite{nayak2005emotional}.

\section{Conclusion}
This paper demonstrates that a motion-controlled robotic arm can inherit the human operator's emotions, then both describes and evaluates an approach for classifying human emotions based on motion-controlled robotic avatar motion behaviours in interactive control scenarios. We extracted the emotion-related features from robot end-effector data and developed a DTW-based algorithm to classify individual subjects' emotions. We further develop an alternative CNN-based algorithm to classify emotions. The training model used could be subject-dependent or independent. Analysis of a dataset of 6000 tasks using a motion-controlled robotic avatar platform found that our approach achieved up to 83.3$\%$ accuracy in recognizing the user's emotion. Our approach is highly suited to motion-based telerobotic use cases when compared to traditional methods. We discuss how this method can be applied to current remote robot operations to build efficient, safe, and human-centred interactions. Furthermore, we explore promising future applications for this approach, including virtual robotic avatars, emotional intelligence encounters between man and machin,e and AI-assisted emotive-motion dampening.


%





\ifCLASSOPTIONcaptionsoff
\newpage
\fi



%
\bibliographystyle{unsrt}
\bibliography{main}

\end{document}